%% file: main.tex
\documentclass[manuscript,screen]{acmart}
\AtBeginDocument{%
  }

\setcopyright{acmlicensed}
\copyrightyear{2018}
\acmYear{2018}
\acmDOI{XXXXXXX.XXXXXXX}
\acmConference[Conference acronym 'XX]{Make sure to enter the correct
  conference title from your rights confirmation email}{June 03--05,
  2018}{Woodstock, NY}
\acmISBN{978-1-4503-XXXX-X/2018/06}

\begin{document}

\title{Towards Data-Centric AI: A Comprehensive Survey of Traditional, Reinforcement, and Generative Approaches for Tabular Data Transformation}


\author{Dongjie Wang}
\affiliation{%
  \institution{University of Kansas}
  \city{Lawrence}
  \country{United States}}
\email{wangdongjie@ku.edu}

\author{Yanyong Huang}
\affiliation{%
  \institution{Southwestern University of Finance and Economics}
  \city{Chengdu}
  \country{China}
}
\email{huangyy@swufe.edu.cn}

\author{Wangyang Ying}
\email{wangyang.ying@asu.edu}
\author{Haoyue Bai}
\email{haoyuebai@asu.edu}
\author{Nanxu Gong}
\email{nanxugong@asu.edu}
\author{Xinyuan Wang}
\email{xwang735@asu.edu}
\author{Sixun Dong}
\email{sixundong@asu.edu}
\affiliation{%
  \institution{Arizona State University}
  \city{Phoenix}
  \country{United States}
}

\author{Tao Zhe}
\affiliation{%
  \institution{University of Kansas}
  \city{Lawrence}
  \country{United States}}
\email{taozhe@ku.edu}

\author{Kunpeng Liu}
\affiliation{%
 \institution{Portland State University}
 \city{Portland}
 \state{Oregon}
 \country{United States}}
 \email{kunpeng@pdx.edu}

\author{Meng Xiao}
\email{shaow@cnic.cn}
\author{Pengfei Wang}
\email{wpf@cnic.cn}
\affiliation{%
  \institution{Computer Network Information Center, Chinese Academy of Sciences}
  \city{Beijing}
  \state{Beijing Shi}
  \country{China}}

\author{Pengyang Wang}
\affiliation{%
  \institution{University of Macau}
  \city{Taipa}
  \state{Macau}
  \country{China}}
\email{pywang@um.edu.mo}

\author{Hui Xiong}
\affiliation{%
  \institution{The Hong Kong University of Science and Technology}
  \city{Guangzhou}
  \country{China}}
\email{xionghui@ust.hk}

\author{Yanjie Fu}
\affiliation{%
  \institution{Arizona State University}
  \city{Phoenix}
  \country{United States}}
\email{yanjie.fu@asu.edu}

\renewcommand{\shortauthors}{Wang et al.}

\begin{abstract}
Tabular data is one of the most widely used formats across industries, driving critical applications in areas such as finance, healthcare, and marketing. 
In the era of data-centric AI, improving data quality and representation has become essential for enhancing model performance, particularly in applications centered around tabular data. 
This survey examines the key aspects of tabular data-centric AI, emphasizing feature selection and feature generation as essential techniques for data space refinement. 
We provide a systematic review of feature selection methods, which identify and retain the most relevant data attributes, and feature generation approaches, which create new features to simplify the capture of complex data patterns.
This survey offers a comprehensive overview of current methodologies through an analysis of recent advancements, practical applications, and the strengths and limitations of these techniques. 
Finally, we outline open challenges and suggest future perspectives to inspire continued innovation in this field.
\end{abstract}




\maketitle

\input{1_introduction}

\input{2_fundamental_tabular}

\input{3_method_feature_selection}

\input{4_method_feature_generation}

\input{5_advanced_fs_fg}

\input{6_comparative_analysis}

\input{7_future_conclusion}

\bibliographystyle{ACM-Reference-Format}
\bibliography{survey,wdj}

\end{document}

%% file: 1_introduction.tex
\section{Introduction}


With the rapid advancement of artificial intelligence (AI), the capabilities of AI models have been extensively explored and developed. 
To further enhance the performance and robustness of AI systems, research focus has increasingly shifted from model-centric AI to data-centric AI. 
High-quality data serves as the foundation for driving innovation and achieving superior model performance.

\begin{figure}[t]
    \centering
\includegraphics[width=0.9\linewidth]{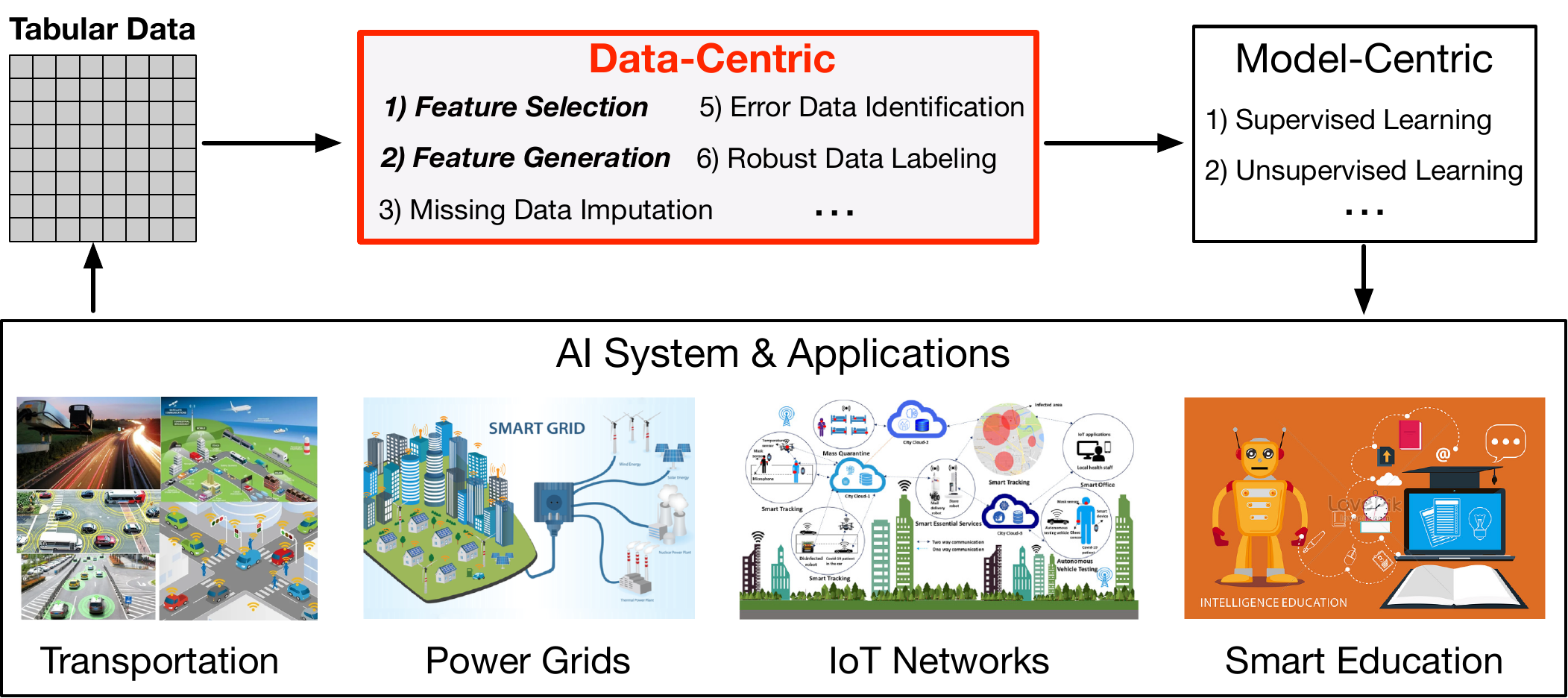}
    \caption{Real-world applications often generate vast amounts of tabular data, making Data-Centric AI essential for optimizing performance. This survey explores the critical aspects of feature selection and generation, key to advancing tabular data-centric AI.}
    \label{fig:survey_focus}
    \vspace{-0.5cm}
\end{figure}

Among various data formats, \textbf{tabular data} remains one of the most popular and critical types, prevalent in fields such as manufacturing, healthcare, marketing, and logistics. Unlike unstructured data, tabular data pose unique challenges in AI, including the curse of dimensionality, complex feature interactions, and feature heterogeneity, as shown in Figure~\ref{fig:survey_focus}. 
Moreover, these domains often require high interpretability and face limitations in data availability compared to fields like computer vision and natural language processing. 
Addressing these challenges is crucial for unlocking the full potential of AI in these traditional areas.

To address these challenges, tabular data transformation plays a crucial role in maximizing the utility of tabular data.
Figure~\ref{fig:overview} reflects the overview of taxonomy of existing transformation techniques.
There are two key tasks in this transformation: \textbf{feature selection} and \textbf{feature generation}. Feature selection focuses on identifying the most relevant and informative features while eliminating redundant ones. In contrast, feature generation involves creating new, meaningful features through mathematical transformations to enhance downstream task performance. Both processes are essential to improve interpretability, predictive accuracy, and efficiency, particularly in domains where high-quality data is limited.

Feature selection methods are broadly categorized into filter, wrapper, and embedded methods~\cite{li2017feature}.
Filter methods rank features based on relevance scores derived from statistical properties of the data, such as correlation with the target variable or mutual information. 
For instance, the K-Best algorithm selects the top-K features based on these criteria. 
Filter methods are computationally efficient and suitable for high-dimensional datasets but often ignore feature dependencies, resulting in suboptimal performance~\cite{mrmr, biesiada2008feature, ding2014identification}.
Wrapper methods evaluate feature subsets iteratively using a predefined machine learning (ML) model. These methods typically outperform filter methods due to their holistic evaluation of feature sets but are computationally expensive, as enumerating all possible subsets is NP-hard ~\cite{gfe,sarlfs, fan2020autofs, fan2021autogfs, altarabichi2023fast}.
Embedded methods integrate feature selection into the model training process by incorporating it as a regularization term in the loss function.
For example, LassoNet combines the sparsity-inducing properties of Lasso with the non-linear modeling capabilities of neural networks, making it effective for handling high-dimensional data with complex relationships among features. 
However, embedded methods are often tailored to specific ML models and may not generalize well to others~\cite{lasso1996,lassonet,rfe,kumagai2022few,koyama2022effective}.
Additionally, hybrid methods combine multiple feature selection strategies and are divided into homogeneous approaches, which integrate techniques of the same type, and heterogeneous approaches, which combine techniques from different categories. The effectiveness of these methods is typically constrained by the limitations of the underlying strategies~\cite{seijo2017testing,pes2017exploiting,haque2016heterogeneous,seijo2019developing}.

Feature generation approaches can be divided into three categories: expansion-reduction, evolution-evaluation, and AutoML-based methods~\cite{chen2021techniques,kusiak2001feature}. 
Expansion-reduction methods first expand the feature space using explicitly defined or greedily selected mathematical transformations and then reduce it by selecting relevant features. 
While computationally efficient, these methods often fail to construct or evaluate complex transformations, resulting in suboptimal performance~\cite{kanter2015deep,khurana2016cognito,lam2017one,horn2019autofeat,katz2016explorekit,dor2012strengthening}. 
Evolution-evaluation methods integrate feature generation and selection into an iterative, closed-loop system optimized by evolutionary algorithms. 
These approaches generate effective features but are computationally expensive and unstable due to reliance on discrete decision-making~\cite{wang2022group,khurana2018feature,tran2016genetic,zhu2022evolutionary,xiao2022traceable}. 
AutoML-based methods treat automated feature generation as an AutoML task, leveraging model architecture search to identify optimal transformations efficiently. 
However, their scalability and robustness are often limited in practice~\cite{chen2019neural,zhu2022difer,elsken2019neural,li2021automl,he2021automl,karmaker2021automl}.

In recent years, advanced techniques such as \textbf{reinforcement learning (RL)} and \textbf{generative AI} have emerged as innovative approaches to address the limitations of traditional feature selection and feature generation methods. 
The trial-and-error nature of RL makes it particularly effective for simulating discrete selection and generation processes, optimizing downstream task performance, and reducing feature redundancy. For instance, Liu et al. formulate feature selection as a multi-agent RL task, where each agent corresponds to a feature, and the action space involves selecting or deselecting features to maximize predictive performance~\cite{liu2019automating}. RL-based frameworks not only automate the exploration of feature spaces but also generate extensive records that encapsulate valuable feature engineering knowledge.
Building on this foundation, generative AI provides a powerful means to encode and reconstruct such feature learning knowledge. 
Inspired by this capability, Wang et al. propose integrating RL and generative AI by using RL-based frameworks as data collectors to automatically generate feature transformation records. These records are then preserved in an embedding space using generative AI, which enables the reconstruction of enhanced feature selection and generation strategies~\cite{NIPS@MOAT,GAINS}. 
This synergy between RL and generative AI creates a novel paradigm for automated data science, offering new insights and opportunities for advancing feature engineering.


To advance the development of tabular data-centric AI and enhance its applicability in traditional domains, this survey offers a comprehensive review of feature selection and generation techniques within this context.
To establish a strong foundation, we discuss the fundamentals of tabular data, emphasizing its structured nature, the distinct challenges it poses compared to other data types, and its critical role in traditional domains. F
or feature selection, we explore both traditional single-view methods, including filter, wrapper, embedded, and hybrid approaches, and multi-view methods encompassing supervised, semi-supervised, and unsupervised techniques. 
Traditional feature generation methods, ranging from manual engineering based on domain expertise to automated techniques, are similarly reviewed. 
Advanced methodologies, such as reinforcement learning (RL) and generative AI, are introduced to illustrate their potential in automating feature engineering processes and addressing limitations of traditional approaches. 
Through a detailed comparative analysis, we identify the strengths and weaknesses of traditional and advanced methods, offering practical guidelines for their application. Finally, we discuss open challenges and outline future research directions, including trends such as AutoML, explainable AI, large language models (LLMs), multi-modality, and privacy-conscious feature engineering with federated learning. 
By synthesizing these perspectives, this survey provides actionable insights for researchers and practitioners, paving the way for continued innovation in tabular data-centric AI.

The survey is organized as follows: \textbf{Section 2} discusses the fundamentals of tabular data, highlighting its structured nature, challenges compared to other data types, and the importance of feature engineering in this context. \textbf{Section 3} reviews traditional feature selection methods, focusing on single-view approaches (filter, wrapper, embedded, and hybrid methods) and multi-view approaches (supervised, semi-supervised, and unsupervised techniques).
\textbf{Section 4} examines traditional feature generation methods, including manual and automated approaches, and their contributions to enhancing model performance.
\textbf{Section 5} explores advanced methods in feature selection and generation, emphasizing RL and generative AI, and their ability to automate and optimize feature engineering processes. 
\textbf{Section 6} presents a comparative analysis of traditional and advanced methods, providing guidelines and best practices for their application. 
\textbf{Section 7} identifies key challenges and outlines future research directions, such as AutoML, explainable AI, LLMs, multi-modality, and federated learning. 
Finally, 
\textbf{Section 8} concludes the survey by summarizing key findings and emphasizing the importance of innovation in feature engineering for tabular data-centric AI.

\noindent\textbf{Compared with existing literature survey.}
Existing surveys on automated feature engineering often focus on general AI tasks or specific domains, leaving a gap in addressing the unique challenges of tabular data.
For instance, Li et al.~\cite{li2017feature} provide a comprehensive overview of feature selection techniques but primarily emphasize traditional methods such as filter, wrapper, and embedded approaches, without delving into recent advancements like reinforcement learning (RL) and generative AI. 
Similarly, surveys by Chandrashekar and Sahin~\cite{chandrashekar2014survey} and Tang et al.~\cite{tang2014feature} focus on feature selection, overlooking the complexities of feature generation. 
On the other hand, studies focusing on feature generation, such as those by Alhassan et al.~\cite{mumuni2024automated} tend to emphasize automated feature engineering for unstructured data, providing limited insights into the structured relationships and interpretability challenges inherent in tabular data. 
Our survey bridges these gaps by integrating discussions on both feature selection and feature generation within a unified framework, categorizing feature selection into single-view (e.g., filter, wrapper, embedded, hybrid) and multi-view (e.g., supervised, semi-supervised, unsupervised) approaches. We systematically explore both traditional and advanced feature generation techniques, including RL and generative AI, and provide actionable guidelines through comparative analysis. Furthermore, we delve into emerging trends such as explainable AI, large language models, multi-modality, and federated learning, offering a forward-looking perspective that extends beyond prior surveys.

%% file: 2_fundamental_tabular.tex
\begin{figure*}[t]
    \centering
\includegraphics[width=1.1\linewidth]{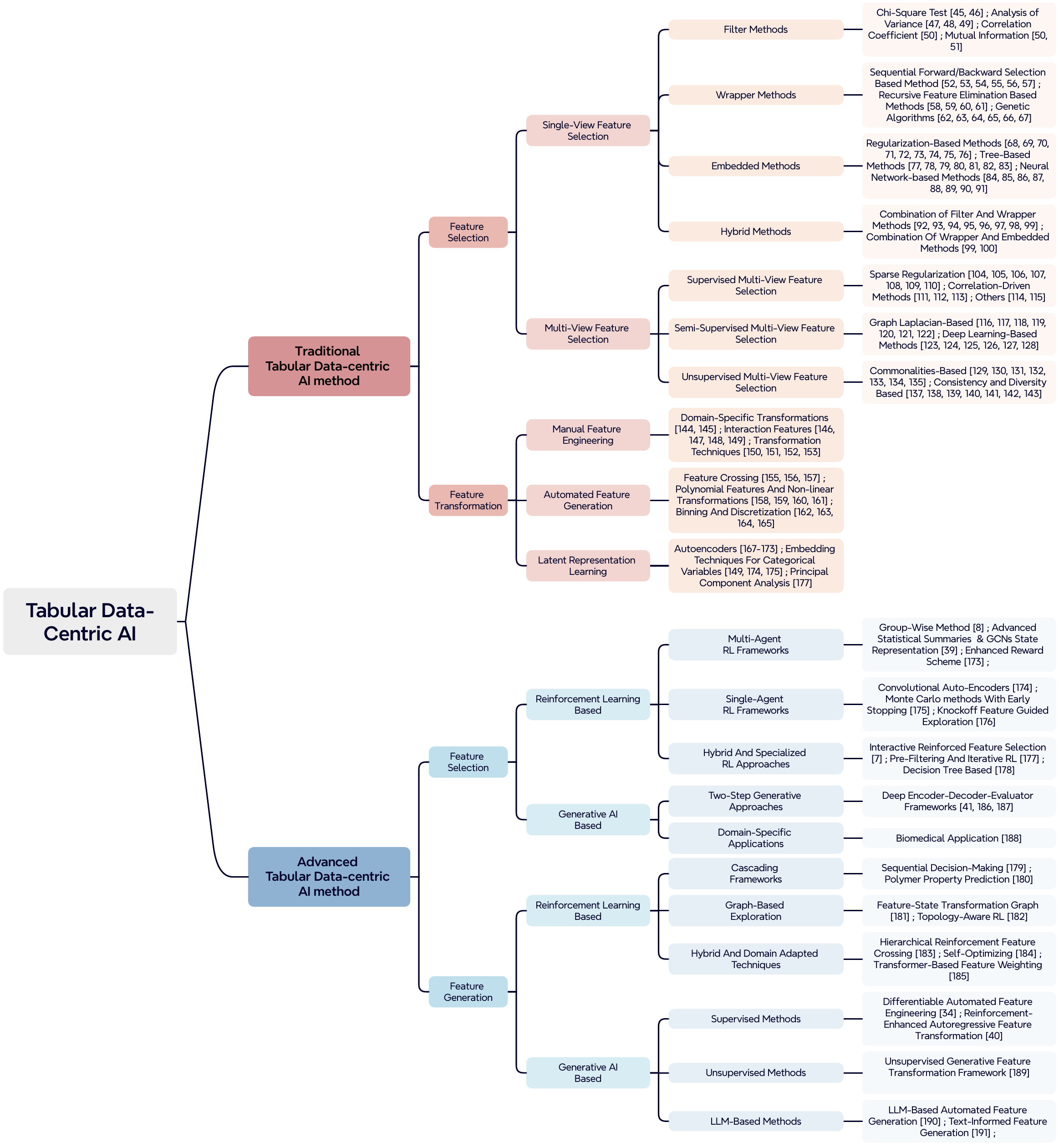}
    \caption{An overview of the taxonomy for existing tabular data-centric AI techniques.}
    \label{fig:overview}
    \vspace{-0.3cm}
\end{figure*}
\vspace{-0.3cm}
\section{Fundamentals of Tabular Data}

Tabular data is widely used in various AI applications and domains, including finance, healthcare, and marketing. 
It is structured into rows (representing instances) and columns (representing features), making it well-suited for organization and manipulation.
However, compared to unstructured data (e.g., images, text), tabular data presents unique challenges, such as heterogeneity (numerical, categorical, ordinal, and temporal data), redundant or irrelevant features, and complex feature interactions, all of which hinder model efficiency and performance.

To enhance data quality and representation, tabular data-centric AI focuses on two core tasks: feature selection and feature generation. These tasks address the aforementioned challenges, creating a refined and distinguishable data space to improve the utility of tabular data in  AI applications.

\textbf{Feature selection} is the process of identifying and preserving the most relevant features while eliminating redundant or irrelevant ones.
Formally, consider a dataset $D = \{(x_i, y_i)\}_{i=1}^n$, where $x_i \in \mathbb{R}^d$ represents the feature vectors for $n$ samples, each comprising $d$ features $\{f_1, f_2, \dots, f_d\}$, and $y_i$ denotes the corresponding target variable. 
The objective of feature selection is to identify a subset $S \subseteq \{1, \dots, d\}$ such that a model trained on the features indexed by $S$ achieves optimal performance while minimizing redundancy among the selected features.

\textbf{Feature generation} is the process of expanding the feature space by constructing new features through mathematical transformations of the original features. The objective is to capture complex relationships within the data while maintaining interpretability, thereby enhancing the performance of downstream tasks. Formally, given a dataset $D = \{(x_i, y_i)\}_{i=1}^n$, feature generation produces a transformed dataset $D' = \{(x'_i, y_i)\}_{i=1}^n$, where $x'_i$ is derived by applying mathematical operations (e.g., addition, subtraction, multiplication) to $x_i$. These newly generated features aim to improve predictive performance by leveraging the enhanced feature representations.

Unlike unstructured data, where feature extraction is largely automated through deep learning architectures (e.g., convolutional neural networks for images or language models for text), tabular data lacks an inherent structure and typically depends on manual feature engineering techniques, such as the creation of interaction terms and aggregations. 
Emerging approaches, including reinforcement learning and generative AI, are paving the way for automating feature engineering in tabular datasets, thereby reducing manual effort and enabling scalable and systematic feature extraction.
By addressing these challenges through feature selection and generation, tabular data-centric AI seeks to transform raw tabular datasets into representations that are optimized for interpretability, computational efficiency, and predictive accuracy. 
This approach bridges the gap between traditional manual methods and automated techniques, unlocking new potential for tabular data analysis.

%% file: 3_method_feature_selection.tex
\section{Traditional Approaches to Feature Selection}
Feature selection, a key preprocessing step in machine learning, reduces dimensionality by identifying informative features and removing redundancy, enhancing efficiency and model performance. Methods are categorized as single-view or multi-view, with single-view approaches classified into filter, wrapper, and embedded techniques, and multi-view methods divided into supervised, semi-supervised, or unsupervised based on label availability. 
In this section, we discuss the advantages and limitations of traditional feature selection methods.

\subsection{Single-view  Feature Selection}

\subsubsection{Filter-based Feature Selection}
Filter-based feature selection identifies features based on their intrinsic properties or statistical relationships with the target label, using metrics such as statistical tests, correlation, or mutual information, and operates independently of specific machine learning models. 

The chi-square test \cite{franke2012chi,mchugh2013chi}, a widely used statistical method, is commonly applied to test the independence of two variables or to evaluate the fit of a sample to a known distribution. This testing method is used in filter-based feature selection to identify features by constructing a contingency table between each feature and the target label, recording feature value frequencies across label categories, and calculating the chi-square statistic or p-value to identify features significantly associated with the target label. However, the chi-square test cannot be directly applied to continuous features, as it requires discretizing the data, potentially leading to information loss. In contrast, analysis of variance (ANOVA)-based feature selection \cite{wang2022ANOVA,lazar2012ANOVA,st1989ANOVA} can directly handle continuous features by comparing between-group and within-group variance. This method evaluates the influence of features on the target label, effectively identifying those significantly associated with it.

To capture feature correlations within the data, filter-based feature selection uses the Pearson correlation coefficient~\cite{zhou2022COVMI} to evaluate the relationships between features and the target label, as well as among features themselves. By eliminating redundancy and identifying the most relevant features, this process ensures effective feature selection. However, the correlation coefficient is limited to capturing linear relationships. To account for nonlinear relationships within the data, a filter-based feature selection approach leveraging mutual information is proposed \cite{battiti1994MI,zhou2022COVMI}. Mutual information-based feature selection assesses the dependency between features and the target variable to identify the most relevant ones, while also reducing redundancy by analyzing mutual information between features. Methods like mRMR (Maximum Relevance Minimum Redundancy)~\cite{Peng2005PAMI} aim to maximize relevance to the target and minimize redundancy among the selected features. Although filter-based feature selection is computationally efficient and model-agnostic, it typically considers features in isolation, overlooking their interactions. Furthermore, its lack of model-specific optimization may restrict its ability to enhance performance.

\subsubsection{Wrapper-based Feature Selection}
To address the limitations of filter-based feature selection methods, Wrapper-based Feature Selection evaluates feature subsets by training machine learning models and assessing performance metrics such as accuracy or F1 score, thereby capturing interactions between features rather than individually evaluating them. Wrapper-based Feature Selection methods are commonly categorized into three types: sequential forward or backward selection, recursive feature elimination, and genetic algorithms.

The first category of wrapper-based methods includes Sequential Forward Selection (SFS) and Sequential Backward Selection (SBS). SFS uses a greedy strategy to iteratively add features to an initially empty set, aiming to identify the subset that maximizes model performance based on a predefined evaluation function~\cite{whitney1971direct}. Conversely, SBS begins with all features and iteratively eliminates redundancies, ultimately identifying the subset that maximizes model performance~\cite{marill1963effectiveness}. SFS is constrained by the nested subset problem, where features selected in earlier iterations cannot be removed, even if they later become redundant as additional features are included. Similarly, SBS cannot reinstate features that were previously removed, which may result in the exclusion of valuable feature combinations. To address these limitations, Pudil et al. proposed Sequential Forward Floating Selection (SFFS) and Sequential Backward Floating Selection (SBFS)~\cite{pudil1994floating}. SFFS combines forward inclusion with backward elimination within a dynamic and flexible framework, iteratively adding performance-enhancing features and removing redundant ones to address the nested subset problem in SFS. Additionally, SBFS begins with the complete feature set, iteratively removing the least significant features while reinstating previously excluded ones if they improve subset performance. Due to the limitations of fixed criteria in SFFS, Somol et al. proposed Adaptive Sequential Forward Floating Selection (ASFFS), which employs a dynamic adjustment mechanism to adapt the inclusion and removal process based on the evolving characteristics of the feature subset. This approach enhances flexibility and improves performance, particularly in complex datasets~\cite{somol1999adaptive}. Building on the SFFS framework, the improved forward floating selection (IFFS) algorithm introduces a novel search step called "weak feature replacement." At each sequential stage, this step removes a feature from the current subset and replaces it with a new one, thereby improving overall performance~\cite{nakariyakul2009improvement}. To alleviate the computational burden of SFS, Ververidis et al. employed a t-test to estimate the required number of cross-validation repetitions. This method leverages the performance of a Bayesian classifier, modeled as a random variable, to reduce computational costs~\cite{ververidis2005sequential}.

The second category of wrapper-based methods is Recursive Feature Elimination (RFE), which iteratively removes the least important features based on their impact on model performance until the desired number of features is selected or performance stabilizes. Guyon et al. integrated Support Vector Machines (SVM) with Recursive Feature Elimination (RFE), forming SVM-RFE, to iteratively identify and remove irrelevant or redundant features based on the maximum margin principle of SVM~\cite{guyon2002gene}. SVM-RFE is prone to "correlation bias" when dealing with highly correlated features, which can lead to underestimating their importance and excluding jointly informative features. To overcome this issue, Yan et al. proposed a Correlation Bias Reduction (CBR) strategy for SVM-RFE, systematically grouping highly correlated features and selecting representative features from each group~\cite{yan2015feature}. You et al. incorporated Partial Least Squares (PLS) into the RFE framework to effectively capture the joint contributions of correlated features and reduce correlation bias. At the same time, they applied a localized recursive approach with One-Versus-All (OVA) decomposition to adapt RFE for multi-class tasks~\cite{you2014feature}. Similarly, Yoon et al. integrated the Minimum Redundancy Maximum Relevance (mRMR) criterion with SVM-RFE to balance feature relevance to the target class and redundancy among features. By employing a dual-ranking mechanism and using average correlation coefficients, they dynamically updated feature rankings during the recursive elimination process~\cite{yoon2009mutual}.

The third category of wrapper-based methods, Genetic Algorithms (GA) for Feature Selection, optimizes feature subsets by simulating natural selection. It iteratively evolves populations through selection, crossover, and mutation to improve model performance. Unlike traditional sequential methods, which struggle with high-dimensional data, Siedlecki et al. proposed a GA-based approach for large-scale feature selection, incorporating genetic operators and a penalty-based fitness function to ensure scalability and robustness~\cite{siedlecki1989note}. However, due to the binary selection mechanism in GA, it cannot capture subtle variations in feature importance. To address this, Raymer et al. proposed a joint framework that integrates feature selection and extraction, employing real-valued feature weights for scaling and finer-grained adjustments~\cite{raymer2000dimensionality}. The Hybrid Genetic Algorithm (HGA) enhances the GA framework by incorporating multiple local search operations. It combines GA's global search capability with the fine-tuning precision of local search and provides flexible control over the size of the selected feature subset~\cite{oh2004hybrid}. Too et al. introduced the Rival Genetic Algorithm (RGA) and its fast variant (FRGA) to improve the efficiency of traditional genetic algorithms. RGA employs a competition strategy to separate chromosomes into winners and losers, facilitating high-quality parent selection. It also utilizes a three-parent crossover with a dynamic mutation rate to balance exploration and exploitation. FRGA extends RGA by incorporating a time-adaptive crossover scheme, which improves computational efficiency while maintaining the quality of feature selection~\cite{too2021new}. Additionally, Tsai et al. proposed a genetic algorithm-based approach for feature and instance selection, aiming to improve classification performance and reduce data dimensionality~\cite{tsai2013genetic}. Building on the HGA framework, Peng et al. utilized conditional mutual information to optimize feature relevance and reduce redundancy. This approach generates compact and discriminative feature subsets~\cite{huang2007hybrid}.

Wrapper-based feature selection is effective in identifying feature subsets tailored to specific models by accounting for feature interactions. However, it is computationally intensive, prone to overfitting, and less suitable for large-scale applications due to its limited generalizability and inefficiency in high-dimensional datasets.

\subsubsection{Embedded-based Feature Selection}
Embedded-based feature selection methods integrate feature selection and model construction into a unified framework by selecting features alongside model optimization during training. Unlike filter-based approaches, these methods capture feature interactions and prioritize those that contribute the most to model performance. Compared to wrapper-based methods, embedded-based methods avoid repeated model training, thereby reducing computational cost. Embedded-based feature selection methods can be broadly classified into three categories: regularization-based methods, tree-based methods, and neural network-based methods.

Regularization-based methods, such as Lasso and Ridge regression, are classical approaches to feature selection that incorporate $\ell_{1}$ and $\ell_{2}$-regularization terms into the objective function, respectively. These methods effectively reduce the coefficients of redundant features to zero (in the case of Lasso) or near zero (in the case of Ridge)~\cite{LASSO_Ridge}. Until now, extensive research has introduced various regularization terms to enhance the performance of feature reduction. In~\cite{L20norm_1, L20norm_2, L20norm_3, L20norm_4, L20_L21}, the $\ell_{2,0}$-norm is adopted to constrain the feature selection matrix, ensuring that the number of non-zero rows matches the desired number of selected features $k$. This approach enables the learning of a sparse selection subspace, achieving feature selection by enforcing model sparsity. The $\ell_{2,0}$-norm is widely recognized as a suitable choice for feature selection tasks. However, solving the $\ell_{2,0}$-norm directly poses significant computational challenges. To address this, the $\ell_{2,1}$-norm is adopted in~\cite{L20_L21, L21} as a computationally efficient alternative for enforcing sparsity. Furthermore, Nie et al.\cite{L2p_norm} proposed the $\ell_{2,p}$-norm as a flexible alternative to the $\ell_{2,0}$-norm, where $p$ is a tunable hyperparameter within the range $0 < p \leq 1$. Additionally, Shi et al.\cite{L2(1-2)} introduced the nonconvex regularizer $\ell_{2,1-2}$, along with an iterative algorithm, to achieve optimal sparse feature selection.

The second category of embedded-based feature selection approaches, tree-based methods, leverages tree models to facilitate efficient feature selection. For instance, a regularization framework is incorporated into random forest models in~\cite{treeRegularization}, which utilizes information gain and imposes penalties on similar features to reduce redundancy. Based on decision trees, FWDT~\cite{FWDT} enhances the ReliefF algorithm by measuring distances between samples and their $k$-nearest neighbors for preliminary feature screening. It then constructs a tree model using the calculated feature weights to identify and exclude redundant features effectively. In~\cite{IRFS}, a closed-loop architecture is proposed, integrating interactive reinforcement learning with decision tree feedback. This approach uses the tree-like feature hierarchy generated by the decision tree to improve state representation and refine the reward mechanism, balancing effectiveness and efficiency in automated feature selection. Inspired by this framework, the flexible neural tree model~\cite{FNT} is constructed using a genetic algorithm. Feature importance is quantified through the evaluation of connection weights and activation function parameters associated with each feature. In~\cite{PStree}, Chen et al. introduced a power set tree that maps feature subsets to tree nodes using an ordered structure. By incorporating search strategies and pruning rules, this method efficiently facilitates feature selection and identifies the optimal or near-optimal feature subset. Additionally, FSFOA~\cite{FSFOA}, an evolutionary algorithm inspired by tree growth processes, reformulates the feature selection problem as a discrete search task. Using classification accuracy as the fitness function, FSFOA effectively eliminates irrelevant and redundant features. Similarly, the iTGA algorithm~\cite{iTGA} simulates the tree growth process, optimizing feature selection by searching for the optimal tree representation of the feature subset within the tree model space.

The third category of embedded feature selection methods, neural network-based approaches, utilizes the structural and functional properties of neural networks to achieve efficient feature selection. In~\cite{FSDNN}, significant features are identified by analyzing the activation of the first layer in deep neural networks, prioritizing those that contribute most to the overall network response. In SDAE-LSTM~\cite{SDAE_LSTM}, a stacked denoising autoencoder is used for feature extraction, while feature selection is conducted by evaluating the mutual information between features and prediction targets, retaining only strongly correlated features. In~\cite{feedforward_NN}, a feedforward neural network incorporates a regularization term into the cross-entropy error function to reduce sensitivity to input variations. Feature selection is performed by evaluating the impact of removing individual features on classification error. To address redundancy among selected features, \cite{NN_GroupLasso1} and \cite{NN_GroupLasso2} integrate neural networks with Group Lasso regularization, introducing group sparsity to prune grouped weights and eliminate non-informative features. Furthermore, to capture nonlinear relationships between data points and their labels, \cite{NFSN, NFSNN} employ nonlinear mapping with row sparsity constraints to derive a discriminative feature subset. These methods utilize $\ell_{2,p}$-norm regularized neural networks with hidden layers, achieving a balance between effective feature selection and minimal redundancy. The aforementioned methods rely on a single neural network for feature selection. However, single neural networks are inherently unstable, as feature importance evaluation can be influenced by initialization parameters. To address this, \cite{ANNs} proposes a robust approach that evaluates feature contributions by analyzing connection weights across multiple trained neural network models. Feature importance rankings are aggregated across these models, and a confidence score is computed to quantify the overall impact of each feature on the output.

\subsubsection{Hybrid Methods for Feature Selection}
Single feature selection techniques, such as filter-based methods, are computationally efficient but often overlook interactions between features. Wrapper-based methods, on the other hand, can capture these interactions but come with high computational costs. Embedded-based methods strike a balance between efficiency and interaction by incorporating feature selection into the model training process. However, their performance is highly sensitive to hyperparameters, such as the regularization strength. Hybrid methods leverage the advantages of filter, wrapper, and embedded approaches, enhancing feature selection accuracy without sacrificing efficiency. By addressing the shortcomings of individual methods, they offer a more robust and effective solution for feature selection. Recent studies classify hybrid methods into two main categories: combinations of filter and wrapper methods, and combinations of wrapper and embedded methods.

The first category of methods in hybrid approaches typically involves two stages: an initial global filtering step to create a candidate feature subset, followed by a wrapper-based optimization to identify the optimal subset. For example, the MIMAGA-Selection method uses mutual information maximization to select an initial feature subset and then applies genetic algorithms to further reduce the dimensionality of this subset~\cite{MIMAGA}. In a similar manner, the studies in ~\cite{EGA} and ~\cite{GALSF} both utilize multiple filter-based methods for preliminary feature screening and ranking, followed by optimization using genetic algorithms to refine the critical feature subset. FG-HFS~\cite{FG-HFS} uses spectral clustering and the approximate Markov Blanket principle to remove redundant features through a filter-based approach. This is followed by a genetic algorithm to search and evaluate feature subsets, selecting the subset with the highest evaluation score. HFSIA~\cite{HFSIA} employs the Fisher filtering method to rank features based on importance, reducing the size of the candidate feature set. It then combines a metaheuristic strategy, utilizing a clone selection algorithm to search the filtered feature set until it identifies the best subset. In~\cite{filter_wrapper}, the initial filtering stage uses information gain and F-score indicators to eliminate redundant and irrelevant features. This is followed by a sequential forward or backward search to iteratively add or remove features until the optimal subset is determined. Similarly, HFS-C-P~\cite{HFS-C-P} removes irrelevant or weakly correlated features using a filter-based method and clusters strongly correlated features into groups. Representative features from each cluster are then selected using a particle swarm optimization (PSO) algorithm to form the final feature subset. In this category of hybrid methods, some algorithms take an inverse approach by first applying a wrapper method to identify a candidate feature subset, followed by a filter method to remove redundant features. The method in ~\cite{HGA} uses a genetic algorithm to perform a global search across the entire feature space to identify a candidate subset. Then, the mutual information criterion is applied to evaluate the contribution of each candidate feature, and an iterative elimination process refines the subset.

The second category of methods in hybrid approaches has also proven to be promising. For instance, Nugroho~\cite{wrapper_RF} combines forward and backward elimination strategies with cross-validation performance evaluation and utilizes random forests to optimize feature subsets iteratively. Likewise, FSPP~\cite{FSPP} evaluates feature importance by leveraging the posterior probability outputs of a multilayer perceptron to calculate each feature's contribution to the scores. Features with the lowest contributions are then removed progressively through a recursive process.

\subsection{Multi-view Feature Selection}
Multi-view data is commonly found in real-world applications, where the same samples are described by heterogeneous features from multiple perspectives~\cite{multi-view-survey}.  When single-view feature selection is applied to multi-view data, all features from different views are simply combined into a single view. However, this approach fails to account for the complementary and consistent information shared across views~\cite{FS-review}. To address this issue, multi-view feature selection methods have been designed to leverage both intra-view and inter-view information, enabling the identification of features that are relevant within individual views and consistent across multiple views~\cite{MFS-review}. Next, we will explore multi-view feature selection methods from the perspectives of supervised, semi-supervised, and unsupervised learning, depending on the availability of label information.

\subsubsection{Supervised Multi-view Feature Selection}
Supervised multi-view feature selection methods can broadly be classified into two categories: sparsity-regularization methods and correlation-driven methods. In the first category of methods, Xiao et al. developed a two-view feature selection method for cross-sensor iris recognition based on $\ell_{2,1}$-norm regularization, effectively addressing sparsity while minimizing misclassification errors~\cite{xiao-MUFS-10}. Lin et al. further advanced sparse regularization by introducing locally sparse constraints combined with block computing techniques, significantly improving computational efficiency for large-scale datasets through the decomposition of optimization problems into multiple smaller sub-problems~\cite{lin-MUFS-6}. Lin et al. also developed a framework combining weighted shared loss and the maximum margin criterion (MMC) to select discriminative features, emphasizing inter-class and intra-class structural information~\cite{lin-MUFS-5}. Cui et al. proposed the Multi-view Stable Feature Selection with Adaptive Optimization of View Weights (MvSFS-AOW), which dynamically adjusts view weights, incorporates unknown data, enhances robustness, and reduces the need for extensive parameter tuning~\cite{cui-MUFS-1}. To enhance the robustness of supervised multi-view feature selection, Lan et al. proposed a framework that combines capped $\ell_{2}$-norm with $\ell_{2,p}$ regularization to effectively address noise and ensure joint sparsity across multiple views~\cite{lan-MUFS-9}. Moreover, the difficulty in solving sparse regularization terms embedded in models has led to the development of several supervised multi-view feature selection methods, leveraging an extended Alternating Direction Method of Multipliers (ADMM) to improve computational efficiency. Lin et al. employed ADMM to design a sharing multi-view feature selection framework, reducing computational complexity and enabling efficient processing of large-scale datasets~\cite{lin-MUFS-7}. Likewise, Men et al. utilized ADMM in a supervised multi-view feature selection framework, incorporating sample-partition and view-partition strategies to compute category-specific loss terms and capture inter-view consistency and distinct characteristics~\cite{men-MUFS-13}.

Compared to the first category of methods, the second category focuses on capturing intra-view and inter-view correlations to better represent or preserve the local and global structures of the data. Jing et al. proposed Intra-view and Inter-view Supervised Correlation Analysis (I²SCA), a method that captures both intra-view and inter-view correlations using a unified objective function. It provides an analytical solution without iterative computation and incorporates a kernelized extension to address non-linear relationships in the feature space~\cite{jing-MUFS-4}. Similarly, Cheng et al. employed hyper-graph regularization to model complex structures in multi-view data. This method combines hyper-graph regularization with low-rank constraints to preserve global data characteristics while enhancing the representation of local relationships~\cite{cheng-MUFS-3}. Xu et al. introduced multi-view scaling support vector machines, which incorporate scaling factors to adjust the contributions of features from different views. This approach emphasizes the importance of weighting features based on their relevance to the classification task~\cite{xu-MUFS-8}. Building on feature integration, Wang et al. proposed a Sparse Multi-modal Learning (SMML) framework tailored for visual tasks. This framework leverages jointly structured sparsity regularizations to integrate heterogeneous visual features, showcasing its effectiveness in both image classification and multi-label tasks~\cite{wang-MUFS-12}. The study by Hao et al.~\cite{hao2024double} introduces a novel splitting structure for observed labels, designed to mitigate noise while retaining both common and complementary information. It proposes a new regularization paradigm that incorporates logic operations to guide the learning direction of variables during the optimization process. Additionally, the approach employs multiplicative update rules with proven convergence to optimize the process effectively. In another work, Hao et al.~\cite{hao2024anchor} present a novel method for reconstructing the global view by integrating information from both shared subspaces and distinct views. This method establishes an accurate mapping between the global view and the label matrix. It employs regularization paradigms focused on extracting information from the candidate view. By enforcing similarity between two latent anchors, the method preserves the candidate view's information in the reconstructed view. This ensures the integrity of shared information while incorporating unique, view-specific details.

\subsubsection{Semi-Supervised Multi-view Feature Selection}
Supervised multi-view feature selection methods typically rely on sufficient labeled data. However, labeling is often a time-consuming and labor-intensive process, which limits the availability of labeled data in practical applications. As a result, semi-supervised multi-view feature selection, which effectively leverages both labeled and unlabeled data, has attracted considerable research interest and shown promising performance. Most existing semi-supervised multi-view feature selection methods are based on graph Laplacians. These methods guide the feature selection process by constructing similarity graphs that exploit the structural consistency between the feature and label spaces, allowing label information to propagate from labeled to unlabeled instances~\cite{iscen2019semi}. Hou et al. proposed a semi-supervised multi-view dimensionality reduction method, which incorporates must-link and cannot-link constraints, represented by Laplacian matrices, to capture both local geometric structures and global relationships among samples. This approach effectively guides the reduction of dimensions using limited labeled data~\cite{hou2010semi}. Shi et al. combined multi-view Laplacian regularization with an $l_{2,1/2}$ norm for semi-supervised sparse feature selection. This method captures local structure through the Laplacian matrix of each view and employs a weighted fusion strategy to leverage complementary information across views~\cite{shi2015semi}. Besides, Shi et al. applied Hessian regularization to encode the local geometric structure of unlabeled data~\cite{shi2016semi}. Jiang et al. proposed a method for learning graph structures across different views to preserve complementary information. This approach fuses view-specific graphs to extract consistency information and dynamically optimizes similarity structures to better utilize unlabeled data~\cite{jiang2022semi}. Zhang et al. constructed a bidirectional graph using anchor points and adaptively learned sample similarities in the weighted feature space. A label propagation strategy was then employed to transfer label information to unlabeled samples, enhancing feature discriminability~\cite{zhang2023semi}. Jiang et al. further extended concept decomposition to multi-view scenarios by integrating limited labeled data via label propagation and using Laplacian graphs to capture local structure and view consistency. This approach significantly improved semi-supervised multi-label feature selection~\cite{jiang2024semi}.

With the advancement of deep learning techniques, they have been increasingly integrated into semi-supervised multi-view feature selection frameworks, facilitating more efficient and effective dimensionality reduction \cite{survey2022semi, survey2020semi}. These methods utilize deep neural networks to extract complex feature representations from large-scale data, effectively leveraging both limited labeled data and abundant unlabeled data within a semi-supervised framework. Noroozi et al. proposed a deep neural network designed to learn low-dimensional feature representations for semi-supervised multi-view data. By minimizing intra-class divergence and maximizing inter-class distance, the proposed approach utilized labeled data to enhance the model's discriminative power~\cite{noroozi2018semi}. Jia et al. integrated deep learning with density clustering to generate pseudo-labels for unlabeled data, thereby improving feature robustness and representation quality. In addition, they introduced orthogonality and adversarial similarity constraints to mitigate feature redundancy~\cite{jia2020semi}. Wu et al. proposed a graph convolutional network (GCN) that incorporates reconstruction error and Laplace embedding. This design enables adjustable Laplace smoothing through flexible graph filters, allowing the model to effectively integrate local and global structural information across views~\cite{wu2023semi}. In a related approach, Xu et al. employed GCN to generate view-specific data representations by combining neighborhood structure and label information. The approach dynamically updated a unified similarity graph to capture the complementarity and consistency of multi-view data~\cite{xu2024semi}.

\subsubsection{Unsupervised Multi-view Feature Selection}
Obtaining labeled data for multi-view datasets is often difficult and sometimes impractical. This limitation has led to significant interest in unsupervised feature selection methods for such data. Since data from different views often share common information, many unsupervised multi-view feature selection methods have been developed to leverage this consistency and identify informative features. Cao et al. proposed a self-representation learning framework to generate multiple cluster structures, which were fused into a consensus cluster structure to provide discriminative information for feature selection~\cite{cao2023consensus}. Tang et al. embedded multi-view feature selection into a non-negative matrix factorization-based clustering model, learning a shared cluster indicator matrix to capture consistent clustering information across views~\cite{tang2018consensus}. Similarly, Bai et al. integrated common similarity graph learning and pseudo-label learning into a sparse linear regression model to enhance feature selection~\cite{bai2020multi}. Liang et al. employed tensor robust principal component analysis to construct noise-free view-specific similarity matrices while adaptively learning a consensus similarity matrix to preserve the local geometric structure of the data~\cite{liang2023multi}. Liu et al. utilized consensus clustering to generate pseudo-labels, which were then used as the response matrix in a sparse feature selection model to identify key features~\cite{liu2018feature}. Wu et al. approached feature selection by leveraging the consistency of both global topological structures and local geometric structures within the data~\cite{wu2023multi}. Zhang et al. proposed an adaptive method to learn sample-view weights for fusing cluster membership matrices, effectively capturing the consensus cluster structure for feature selection~\cite{zhang2024scalable}.

The aforementioned methods primarily focus on exploring commonalities across views during the feature selection process. But, beyond shared information, each view also exhibits unique characteristics. Thus, many unsupervised multi-view feature selection methods aim to exploit both the consistency and diversity of multi-view data to enhance feature selection. Zhang employed the fuzzy c-means clustering algorithm~\cite{bezdek1984fcm} to learn membership matrices for each view. These matrices were then collaboratively fused with bipartite graphs to explore the consensus cluster structure and local geometric structure~\cite{zhang2024efficient}. Yuan et al. constructed individual similarity graphs for each view to capture view-specific information and imposed a tensor low-rank regularization constraint on the graph tensor to maintain cross-view consistency~\cite{yuan2022multi}. Fang et al. formulated multi-view unsupervised feature selection as an orthogonal decomposition problem, where target matrices were decomposed into view-specific basis matrices and a view-consistent cluster indicator to capture both consensus and diverse information across views~\cite{fang2023joint}. Zhou et al. imposed a nuclear norm on the representation matrix to extract consensus information from different views and introduced the Hilbert–Schmidt Independence Criterion (HSIC) to promote view exclusivity~\cite{zhou2024consistency}. Additionally, Cao et al. proposed constructing multiple mutually exclusive graphs to capture the complementarity of different views. These graphs were then used to learn a view-shared clustering indicator matrix to ensure consistency across views~\cite{cao2024multi}. Tang et al. introduced the CRV-DCL method, which projects data from each view into a relaxed label space comprising common and diverse components. This approach enables the extraction of both shared and unique information for feature selection~\cite{tang2019cross}. As an extension of CRV-DCL, CvLP-DCL developed a cross-view similarity graph learning model with adaptive view weights to preserve the local manifold structure of data, thereby facilitating the selection of discriminative features~\cite{tang2021cross}. Furthermore, to address the challenge of incomplete multi-view data hindering the accurate construction of similarity graphs for each view, Huang et al. proposed to adaptively learn both shared and view-specific similarity matrices by using complementary and consistent information across views to guide feature selection~\cite{Huang2024}.

\subsection{Challenges and Limitations in Traditional Methods}
Feature selection has proven to be highly effective in data preprocessing and dimensionality reduction, highlighting its essential role in machine learning and data mining. In recent years, a variety of single-view and multi-view feature selection methods have been developed. However, several key challenges in feature selection still need to be resolved.
First, collected data often suffers from low quality, including incompleteness, noise, and sparse labeling. Designing robust methodologies to mitigate the impact of noise and outliers, while enabling effective feature selection in such data, remains a key research challenge.
Moreover, in dynamic open environments, collected data frequently exhibits multi-dimensional changes at the sample, feature, and label levels. Developing feature selection models that can efficiently adapt to incremental data while preserving essential knowledge to mitigate catastrophic forgetting is a significant challenge.
Furthermore, variations in view quality caused by dynamic environments, along with decision-making biases such as selection bias and label bias inherent in the original data, present another challenge. Developing trustworthy dynamic feature selection methods that balance efficiency, low redundancy, and fairness is crucial for addressing these issues.

%% file: 4_method_feature_generation.tex
\section{Traditional Approaches to Feature Generation}

Feature generation has been a foundational aspect of machine learning and data science, acting as a critical intermediary between raw data and effective predictive models.
By deriving new features from existing ones using mathematical transformations, practitioners aim to uncover patterns, relationships, and latent insights that are often obscured in the raw data. 
These transformations enhance the data space, leading to improvements in model accuracy, interpretability, and generalization.
This section explores the core methodologies underlying traditional feature generation, examining their strengths, limitations, and their role in shaping the evolution of data-centric AI.

\subsection{Human-Driven Feature Generation}

The feature generation process has traditionally relied heavily on the creativity and expertise of domain specialists. 
This approach involves applying transformations and aggregations to construct features that capture meaningful relationships within the data. 
The following subsections introduce various aspects of human-driven feature engineering, highlighting the methodologies and their impact on data representation.

\subsubsection{Mathematical Transformations} Mathematical operations, such as addition, logarithmic transformations, multiplication, and square root calculations, are frequently employed to derive interaction features. These transformations are designed to capture complex relationships between features, thereby enhancing the expressive power of the feature space~\cite{tukey1977exploratory,osborne2003notes}.
For instance, logarithmic transformations compress data ranges, reducing the influence of extreme values, and are extensively applied in financial data for highly skewed variables like income or sales~\cite{weeks2002introductory}. Taking the logarithm of annual income, for example, stabilizes variance and produces a distribution closer to normal, facilitating more robust model training. Similarly, square root transformations are effective for positively skewed data. In ecological studies, square root transformations are commonly used to normalize species abundance data, enabling improved analysis of biodiversity metrics~\cite{baltosser1996biostatistical}.
In addition to individual transformations, interaction features are often created by combining original features through operations such as multiplication, addition, or ratios~\cite{calder2003feature,apel2013exploring,cameron1993feature}. These interactions capture feature-feature relationships and reveal nonlinear dependencies that may not be detectable by linear models. This approach has been widely adopted in various applications. For example, in recommendation systems, user behavior features (e.g., click rate) are combined with product features (e.g., price) to study interactions between users and products, ultimately optimizing personalized recommendations~\cite{cheng2016wide}.
These mathematical transformations refine data by capturing complex nonlinear patterns, enabling deeper insights and enhancing the expressive power of feature spaces.

\subsubsection{Statistical Representations}
Descriptive statistics, including mean, variance, and skewness, are foundational tools in feature generation. They provide concise summaries of data distributions, reveal critical patterns, and serve as the basis for more advanced analyses. Aggregations across groups further enhance this capability by capturing essential relationships in time-series and grouped data, making these statistical representations highly versatile and effective~\cite{hyndman2018forecasting, bishop2006pattern}.
For example, the mean (average) is widely used as a measure of central tendency across domains such as finance and healthcare. In financial portfolio analysis, the mean daily return of a stock is a key indicator of performance~\cite{sharpe1994sharpe}, while in healthcare, the average length of hospital stays provides insights into operational efficiency~\cite{indicators2019health}. The median, by contrast, is robust to outliers, making it particularly useful in real estate to determine typical housing prices within a neighborhood~\cite{fisher2022state}.
Metrics that quantify variability, such as variance and standard deviation, are critical in understanding data dispersion. In portfolio management, variance is central to balancing risk and reward by measuring the variability of asset returns~\cite{fabozzi2011portfolio}. Similarly, in sensor networks, standard deviation helps identify inconsistencies in temperature readings, enabling effective fault detection~\cite{muhammed2017analysis}.
Higher-order statistics like skewness and kurtosis provide additional insights into data distributions. Skewness captures asymmetry and is particularly useful for studying rare events. In climatology, for instance, it reveals the frequency of extreme temperature occurrences~\cite{rahmstorf2011increase}. Kurtosis, which measures the "tailedness" of a distribution, is widely applied in finance to detect fat-tailed risks, such as those associated with stock market crashes~\cite{nicholas2008black}.
These statistical representations provide concise and interpretable summaries of complex datasets, making them essential for feature engineering and machine learning.

\subsubsection{Domain Knowledge Integration} 
Integrating domain knowledge into feature generation is crucial for enhancing the performance of machine learning models. Unlike generic mathematical or statistical transformations, domain-specific feature generation leverages insights and expertise from specific fields to design features that capture meaningful patterns, relationships, and metrics. This tailored approach ensures that the generated feature space closely aligns with real-world observations and task-specific requirements, often resulting in superior model accuracy and interpretability.
Numerous examples illustrate how incorporating domain knowledge significantly enhances both data understanding and AI performance. In finance, the debt-to-income ratio is a widely used metric in credit risk modeling, measuring an individual's financial stability by dividing total debt by gross income~\cite{tsay2005analysis}. Such domain-specific features provide actionable insights into economic trends and individual risk profiles.
In healthcare, domain knowledge drives the creation of features that reflect patient health and medical risks. For example, the Body Mass Index (BMI), derived from weight and height, categorizes individuals into weight classes and assesses related health risks~\cite{harrell2001regression}. 
In e-commerce, domain-specific features capture user behavior, product attributes, and transactional patterns. For instance, purchase frequency measures how often a customer makes purchases, serving as an indicator of loyalty and engagement~\cite{linden2003amazon}. Another key metric, Customer Lifetime Value (CLV), estimates the total revenue a customer will generate for a business over time, enabling businesses to prioritize customer retention strategies~\cite{fader2005note}.
Domain-specific knowledge is vital for crafting tailored features that elevate model performance and interpretability, uncovering insights that align seamlessly with real-world goals and applications.

\subsection{Automated Feature Generation}

With advancements in technology, automated feature generation has become a focal point for many researchers. The central goal of this field is to replicate and enhance the feature generation process traditionally performed by human experts. By systematically uncovering complex feature interactions, modeling non-linear relationships, and iteratively refining the feature space, these techniques aim to optimize performance for downstream tasks. This section explores the key technical pillars of automated feature generation: feature interaction modeling, non-linear transformation and simplification, and iterative refinement and optimization.

\subsubsection{Feature Interaction Modeling}

Feature interaction modeling aims to identify and leverage complex relationships among features to produce new and effective ones.
A common approach is \textbf{feature crossing}, where new features are generated by combining pairs or groups of existing features. 
For instance, a cross feature \( x_{ij} = f(x_i, x_j) \) might be created using operations such as addition, multiplication, or concatenation, depending on the nature of the data set and the characteristics of downstream task.
In prior literature,
AutoCross~\cite{luo2019autocross} automates cross-feature generation using a beam search algorithm guided by information gain to evaluate candidate feature sets. To address the computational cost of direct evaluations, it employs field-wise logistic regression and successive mini-batch gradient descent, enabling efficient performance estimation. These innovations make AutoCross scalable and practical for large-scale tabular data applications.
Additionally, deep learning methods are used to further integrate interaction modeling into model architectures. Deep \& Cross Networks (DCN)~\cite{wang2017deep} introduce explicit cross-feature layers:
$
x^{(l+1)} = W_l (x^{(l)} \otimes x^{(0)}) + b_l + x^{(l)},
$
where \( x^{(l)} \) is the feature vector at layer \( l \), \( \otimes \) represents outer product operations capturing cross terms, and \( x^{(0)} \) is the original feature vector. 
This formulation allows DCNs to directly model meaningful interactions during training.
Neural Factorization Machines (NFMs)~\cite{he2017neural} combine deep learning with factorization principles to model interactions implicitly.
They utilize low-dimensional latent representations \( v_i \) for each feature \( i \):
\(
y = \sum_{i<j} (v_i^\top v_j) \cdot x_i x_j,
\)
enabling efficient computation of feature interactions without explicitly enumerating all combinations. 

\subsubsection{Non-linear Transformation and Simplification}
In real-world applications, capturing non-linear relationships among features is essential for improving predictive performance. Linear transformations often fail to model the complexity of such relationships, prompting researchers to develop automated frameworks that integrate non-linear transformations. These frameworks enhance the expressiveness and flexibility of feature spaces, enabling models to better capture intricate patterns in the data.
One common approach involves constructing polynomial features, as demonstrated by Polynomial Networks~\cite{livni2014computational}. These networks automatically generate polynomial features from a dataset with features \( x_1, x_2, \ldots, x_n \) up to a specified degree \( k \). The generated features take the form:
\(
\phi_k(x) = \{x_1^{d_1} x_2^{d_2} \cdots x_n^{d_n} \mid \sum_{i=1}^n d_i \leq k\}.
\)
By introducing higher-order interactions among variables, these features significantly expand the model’s ability to capture complex relationships.
Another technique, Kernel Learning~\cite{cortes2009learning}, maps features into higher-dimensional spaces using kernel functions. For example, the radial basis function (RBF) kernel is defined as:
\(
K(x, x') = \exp\left(-\frac{\|x - x'\|^2}{2\sigma^2}\right),
\)
where \( \sigma \) determines the smoothness of the kernel. This transformation effectively separates non-linearly separable data in the original feature space, enabling models to learn more accurate decision boundaries.
Building on these ideas, Deep Polynomial Networks (DPNs)~\cite{chrysos2021deep} combine polynomial transformations with the representational power of deep learning. These networks incorporate polynomial transformations into their architecture as layers, expressed as:
\(
z^{(l)} = \sum_{i=1}^n w_i^{(l)} \cdot \phi_k(x),
\)
where \( \phi_k(x) \) represents polynomial transformations of the input \( x \). By leveraging the depth of neural networks, DPNs can model highly complex patterns, providing a seamless integration of non-linear transformations within a deep learning framework.
In addition to these structured transformations, AutoLearn~\cite{kaul2017autolearn} adopts a regression-based approach to uncover non-linear relationships. By fitting pairwise regression models \( f(x_i, x_j) \) between features \( x_i \) and \( x_j \), AutoLearn generates new features that encode intricate interactions. This automated process enables the discovery of meaningful relationships without requiring domain-specific expertise, improving predictive performance in a scalable and efficient manner.

\subsubsection{Iterative Refinement and Optimization}
When human experts perform feature generation, they typically engage in an iterative process of refining the feature space based on feedback from downstream tasks. Inspired by this workflow, many researchers adopt iterative methodologies to construct and refine high-quality features. Unlike one-time transformations, these approaches employ a cyclical strategy that involves generating, evaluating, and selecting features. 
Feedback from each iteration informs subsequent rounds, facilitating a systematic exploration of the extensive feature space while simultaneously managing feature complexity and reducing redundancy.

The iterative refinement process typically consists of three core steps. First, at each iteration \( t \), a set of candidate features \( \tilde{F}^{(t)} \) is derived from the current feature set \( F^{(t)} \) using a set of transformations \( \mathcal{T} \). This can be expressed as:  
\[
\tilde{F}^{(t)} = \{ \mathcal{T}_i(f) \mid f \in F^{(t)}, \mathcal{T}_i \in \mathcal{T} \},
\]
where \( \mathcal{T}_i \) represents a specific transformation, such as arithmetic operations, polynomial expansions, or domain-specific functions.
Second, each candidate feature \( \tilde{f} \in \tilde{F}^{(t)} \) is evaluated to determine its contribution to the learning task using a metric \( \mathcal{E}(\cdot) \). For example:  
\[
\mathcal{E}(\tilde{f}) = \mathbb{E}_{(x, y) \sim D} [ \text{Perf}(y, \hat{y}(\tilde{f})) ],
\]
where \( D \) is the dataset, \( \hat{y}(\tilde{f}) \) represents predictions using \( \tilde{f} \), and \( \text{Perf}(\cdot) \) is a performance measure (e.g., accuracy or AUC).
Finally, features meeting a predefined performance threshold \( \tau \) are added to the feature set for the next iteration:  
\[
F^{(t+1)} = F^{(t)} \cup \{ \tilde{f} \in \tilde{F}^{(t)} \mid \mathcal{E}(\tilde{f}) \geq \tau \}.
\]
This ensures that only valuable features contribute to subsequent rounds of refinement. The process continues until a stopping condition, such as performance convergence, computational limits, or resource constraints, is satisfied.

Several studies exemplify the strengths of iterative refinement in feature engineering. Krawiec~\cite{krawiec2002genetic} introduced a genetic programming-based framework where feature representations are iteratively refined through evolutionary operators, such as mutation and crossover. These methods evolve features across generations, retaining only those with proven predictive value. Katz et al.~\cite{katz2016explorekit} developed ExploreKit, a framework that systematically applies a rich library of transformations and uses meta-learning to estimate the utility of candidate features. Features with low utility are pruned after evaluation, ensuring efficient use of computational resources. Luo et al.~\cite{luo2019autocross} proposed AutoCross, an iterative framework for large-scale tabular data, which identifies and evaluates high-value cross-features using heuristic and greedy search, discarding suboptimal candidates in each iteration. Tools like One Button Machine (OBM)~\cite{lam2017one} and AutoFeat~\cite{horn2019autofeat} leverage iterative pipelines to repeatedly generate, validate, and retain candidate features. These frameworks focus on maintaining consistent performance gains while balancing computational cost.

Iterative refinement pipelines effectively balance the exploration of the feature space with computational and interpretive efficiency. By incorporating performance feedback at each stage, these frameworks adaptively optimize feature sets, uncovering intricate relationships that simpler methods might fail to detect. Additionally, features such as complexity control and stopping criteria ensure that the resulting feature sets are both impactful for downstream tasks and manageable in terms of computational cost and interpretability. This feedback-driven methodology has become a cornerstone of modern feature engineering, particularly when applied to large and heterogeneous datasets.

\subsection{Challenges and Limitations}
Traditional and automated feature generation methods, while effective in certain contexts, face several critical challenges that limit their broader applicability. First, the iterative nature of feature generation pipelines is inherently time-consuming. 
These methods often involve repeated cycles of transformation, evaluation, and refinement, which demand substantial computational resources, particularly when applied to large and complex datasets. This extensive runtime poses significant barriers for scenarios requiring rapid analysis or real-time decision-making.
Second, these approaches exhibit limited transferability. Feature generation processes are typically tailored to specific datasets or tasks, relying heavily on domain-specific knowledge and customized heuristics. Consequently, the features generated through such methods often fail to generalize effectively to new datasets or tasks, necessitating substantial manual intervention and redesign to adapt to different contexts.
Third, traditional techniques struggle to capture non-linear relationships within data effectively. Many established methods are built around linear transformations or interactions, which are insufficient to model complex, non-linear dependencies. While advanced techniques like polynomial or kernel transformations offer some improvement, they remain constrained by the growing complexity and dimensionality of non-linear relationships, particularly in heterogeneous datasets where dependencies can vary widely. These challenges underscore the limitations of conventional feature generation methodologies and highlight the need for innovative approaches to address these issues.

%% file: 5_advanced_fs_fg.tex
\section{Advanced Methods in Feature Selection and Generation: Reinforcement Learning (RL) and Generative AI}

With the growing success and widespread adoption of Reinforcement Learning (RL) and Generative AI, researchers are increasingly exploring innovative ways to reformulate tabular data-centric AI tasks. The inherent nature of RL aligns seamlessly with the iterative workflows of human experts, as it leverages reward mechanisms to efficiently navigate and optimize the search for improved feature spaces. Meanwhile, Generative AI, renowned for its ability to encapsulate knowledge within latent spaces, offers a promising approach to compress feature knowledge and systematically search for enhanced feature representations. In the following sections, we delve into the advancements in RL and Generative AI for feature selection and generation, highlighting their transformative impact on data-centric AI.

\subsection{RL for tabular data-centric AI}
\noindent\textbf{Why is RL suitable for reformulating tabular data-centric AI?} RL offers a unified framework for automating feature selection and feature generation, enabling a dynamic and adaptive exploration of the feature space. Both tasks can be naturally formulated as Markov Decision Processes (MDPs), allowing RL agents to iteratively refine feature subsets or generate new features with the goal of optimizing downstream performance. 

\begin{figure}[t]
    \centering
\includegraphics[width=0.9\linewidth]{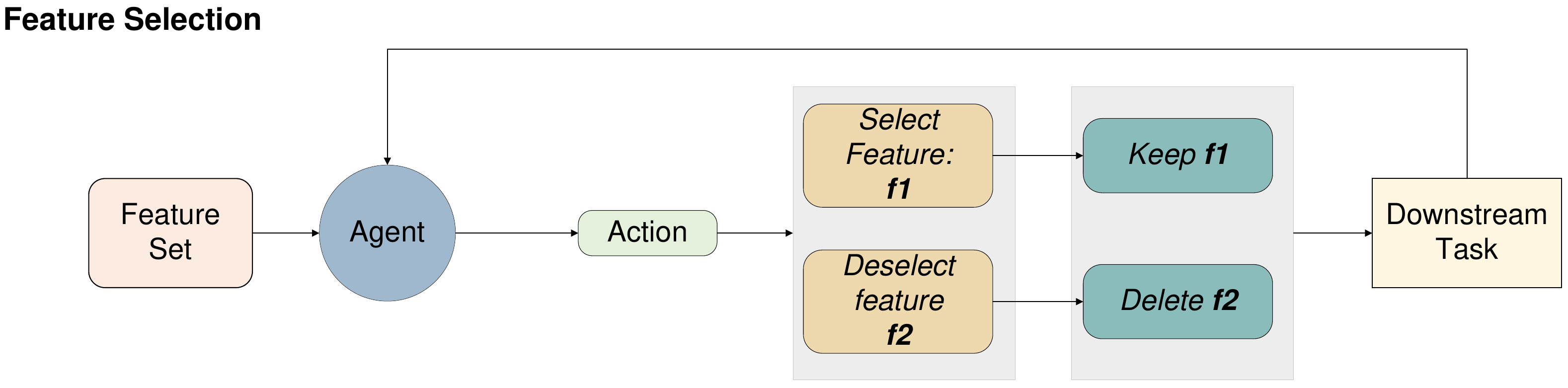}
    \caption{Feature selection is framed as a RL problem, wherein the agent's actions correspond to the selection of individual features. The optimization objective is to enhance performance on downstream tasks.}
    \label{fig:rl_fs}
\end{figure}

\begin{figure}[t]
    \centering
\includegraphics[width=0.9\linewidth]{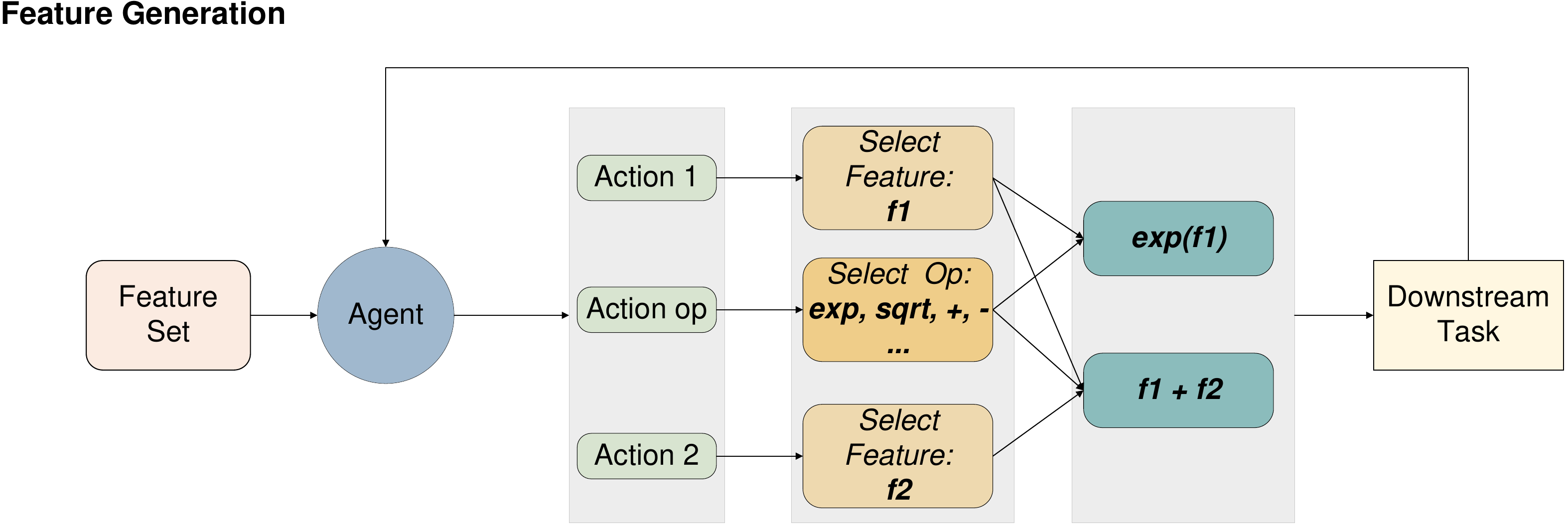}
    \caption{Feature generation is formulated as a RL problem, where the agent's actions involve selecting individual features and applying mathematical operations. The optimization goal is same as feature selection.}
    \label{fig:rl_fg}
    \vspace{-0.3cm}
\end{figure}

Figure~\ref{fig:rl_fs} and Figure~\ref{fig:rl_fg} illustrates the RL formulations for feature selection and feature generation.
Formally,
at iteration \( t \), the \textbf{state} \( S^{(t)} \) represents the current feature set \( F^{(t)} \) and any additional information required for decision-making. For feature selection, \( S^{(t)} = F^{(t)} \); for feature generation, the state includes the transformation history, \( S^{(t)} = \{ F^{(t)}, H^{(t)} \} \), where \( H^{(t)} \) tracks applied transformations.
The \textbf{action} \( A^{(t)} \) determines the operation performed at each iteration. For feature selection, \( A^{(t)} = \{ \text{select } f, \text{deselect } f \mid f \in F^{(t)} \} \); for feature generation, actions involve selecting features and applying mathematical operations: 
\[
A^{(t)} = \{ (f_i, f_j, \mathcal{T}_k) \mid f_i, f_j \in F^{(t)}, \mathcal{T}_k \in \mathcal{T} \},
\]
where \( \mathcal{T}_k \) is a specific transformation (e.g., addition, multiplication).
The \textbf{reward} \( R^{(t)} \) evaluates the quality of the action. For feature selection, the reward is 
\[ R^{(t)} = \text{Perf}(y, \hat{y}(F^{(t+1)})) - \lambda \cdot |F^{(t+1)}| \]
, where \( \text{Perf}(\cdot) \) measures model performance (e.g., accuracy), \( \hat{y}(\cdot) \) represents predictions, and \( \lambda \) penalizes feature set size; for feature generation, the reward also considers novelty and complexity:
\[
R^{(t)} = \text{Perf}(y, \hat{y}(F^{(t+1)})) + \alpha \cdot \text{Novelty}(\tilde{f}) - \lambda \cdot \text{Complexity}(\tilde{f}),
\]
where \( \text{Novelty}(\tilde{f}) \) quantifies the uniqueness of the generated feature \( \tilde{f} \).
The \textbf{state transition} \( S^{(t+1)} \) is determined by the action \( A^{(t)} \). For feature selection, the updated state is
\[ S^{(t+1)} = F^{(t)} \cup \{ f \in F^{(t)} \mid A^{(t)} = \text{select } f \} \]; 
for feature generation, the next state includes the newly generated features and updates the transformation history:
\[
S^{(t+1)} = F^{(t)} \cup \tilde{F}^{(t)}, \quad \tilde{F}^{(t)} = \{ \mathcal{T}_k(f_i, f_j) \mid (f_i, f_j, \mathcal{T}_k) \in A^{(t)} \}.
\]
The \textbf{policy} \( \pi(A^{(t)} \mid S^{(t)}) \) governs the action taken at each state. The objective of the RL agent is to learn an optimal policy \( \pi^* \) that maximizes the cumulative discounted reward:
\[
\pi^* = \arg\max_{\pi} \mathbb{E}\left[ \sum_{t=0}^\infty \gamma^t R^{(t)} \right],
\]
where \( \gamma \) is the discount factor balancing immediate and future rewards.

\subsubsection{RL for Feature Selection}
RL provides a structured approach to feature generation by leveraging the MDP framework. By iteratively generating and refining features, RL overcomes challenges like complex transformations, high-dimensional feature spaces, and intricate interactions, offering an effective solution for automating the feature generation process.

Existing works can be categorized into three technical approaches: \textbf{Multi-Agent RL Frameworks}, \textbf{Single-Agent RL Frameworks}, and \textbf{Hybrid and Specialized RL Approaches}. Multi-agent frameworks emphasize parallel exploration, where tasks are distributed across multiple agents for scalability and coordination. Liu et al. \cite{liu2019automating} introduced a multi-agent RL framework, treating each feature as an independent agent, with advanced state representations such as statistical summaries and graph convolutional networks (GCNs) to enhance decision-making. Building on this, Liu et al. \cite{liu2021automated} employed GCNs to explicitly model relationships between features, improving collaboration among agents through enhanced reward schemes. To reduce computational complexity, Fan et al. \cite{fan2021autogfs} clustered similar features into groups, with each group managed by a single agent, maintaining scalability while reducing resource requirements.

While multi-agent frameworks effectively handle large feature spaces, they can be computationally expensive. Single-agent RL frameworks address this by consolidating decision-making into a single agent, enabling sequential exploration of features. Zhao et al. \cite{zhao2020simplifying} proposed a simplified single-agent RL framework, leveraging convolutional auto-encoders to improve state representation and reduce computational overhead. Liu et al. \cite{liu2021efficient} further enhanced efficiency using Monte Carlo methods with early stopping, minimizing training time. Additionally, Wang et al. \cite{wang2024knockoff} introduced knockoff features to guide exploration, enabling robust feature selection without labeled data.

To further enhance adaptability, researchers have developed hybrid and specialized RL approaches, combining RL with other techniques or adapting it to domain-specific requirements. Fan et al. \cite{fan2020autofs} introduced the Interactive Reinforced Feature Selection (IRFS) framework, which integrates external trainers to guide RL agents with diverse strategies, improving exploration breadth. 
Meng et al. \cite{xiao2025} integrated a multi-agent reinforcement learning system for gene panel selection, using an expert knowledge-guided reward function for label-free biomarker identification. 
Fan et al. \cite{fan2021interactive} incorporated decision tree feedback into RL, leveraging feature hierarchies to improve state representation and personalize agent rewards.

In summary, RL-based feature selection frameworks offer innovative solutions for addressing challenges in high-dimensional and complex feature spaces. By leveraging multi-agent coordination, single-agent optimization, and hybrid strategies, these approaches demonstrate the versatility of RL in advancing feature selection methodologies and improving machine learning model performance.

\subsubsection{RL for Feature Generation}

Feature generation aims to refine or create new features, enhancing the robustness and informativeness of feature spaces to improve downstream machine learning performance. RL has been utilized to automate and optimize this process, providing significant advancements over traditional feature engineering methods. 

Existing works can be categorized into \textbf{cascading frameworks}, \textbf{graph-based exploration}, and \textbf{hybrid and domain-adapted techniques}.
Cascading RL frameworks focus on iterative feature generation through sequential decision-making, leveraging reinforcement learning to build features progressively. Wang et al. \cite{wang2022group} proposed a cascading framework with three agents: two feature agents for selecting candidate features and one operation agent for choosing mathematical operations. This approach iteratively performs feature crossing, generating new features to expand the feature space effectively. Hu et al. \cite{hu2024reinforcement} extended this concept to polymer property prediction, using cascading RL strategies to create meaningful descriptors that improve the explainability and predictive accuracy of models, even when working with low-quality datasets.

Graph-based methods leverage the structural relationships between features to optimize feature generation. Huang et al. \cite{huang2024enhancing} introduced a graph-based reinforced exploration strategy that utilizes a feature-state transformation graph. This approach tracks and reuses valuable transformations, improving adaptability and robustness, particularly for tabular data. Ying et al. \cite{ying2024topology} further advanced this concept for graph data, employing topology-aware RL to reconstruct feature spaces by extracting core subgraphs and encoding topological features using graph neural networks. This ensures the generated features capture critical structural information, enhancing generalization in graph-based machine learning tasks.

Hybrid approaches combine RL with other techniques or adapt it to specific domain challenges, enhancing feature generation's flexibility and effectiveness. Ying et al. \cite{ying2023self} proposed a hierarchical reinforcement feature crossing method to tackle feature interaction challenges. This three-step process—comprising feature discretization, hashing, and descriptive summarization—improves the quality of generated features by creating a discriminative representation space. Xiao et al. \cite{xiao2022self} introduced a self-optimizing RL framework that resolves Q-value overestimation and incorporates advanced state representations, resulting in enhanced policy learning and better feature transformation outcomes. Zhang et al. \cite{zhang2024tfwt} developed a Transformer-based feature weighting method, integrating RL to fine-tune feature weights dynamically based on downstream task feedback. This framework reduces data redundancy and classification variance by capturing feature dependencies through the Transformer’s attention mechanism.

In summary, RL-based feature generation frameworks redefine how features are created and refined, overcoming complexities in high-dimensional spaces. By harnessing cascading strategies, structural insights, and hybrid innovations, they illustrate RL's transformative potential to revolutionize the feature generation task.

\subsection{Generative AI for Tabular Data-Centric AI}

\noindent\textbf{Why is Generative AI suitable for reformulating tabular data-centric AI?} Generative AI has achieved remarkable success in various domains, exemplified by ChatGPT in natural language processing and Stable Diffusion in image synthesis. 
These advancements demonstrate its ability to compress discrete human knowledge into a structured embedding space and reconstruct that knowledge with high fidelity.
Furthermore, generative AI reflects the capability of transferring learned knowledge to different domains and tasks while enabling efficient optimization for better results in the continuous embedding space.
Motivated by these capabilities, researchers have begun exploring the application of generative AI to feature learning in tabular data-centric AI.
Using generative AI offers a promising way to encode, optimize, and transfer feature learning expertise across tasks and domains.

\begin{figure}[t]
    \centering
\includegraphics[width=0.9\linewidth]{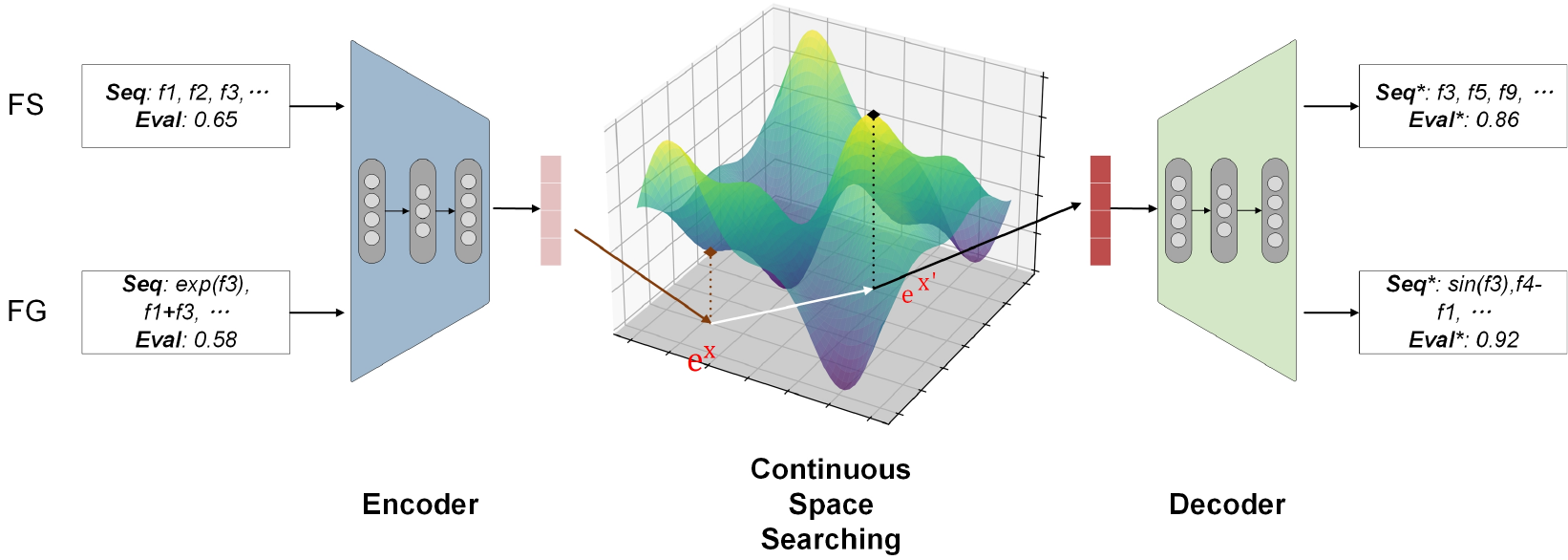}
    \caption{RL extensively explores the feature space, generating abundant feature learning (feature selection/generation) knowledge. Generative AI captures this knowledge in a continuous embedding space, enabling gradient-based search to identify optimized feature spaces.}
    \label{fig:generative AI}
    \vspace{-0.3cm}
\end{figure}

As shown in Figure~\ref{fig:generative AI}, the formulation of generative AI for tabular data-centric AI is structured around an encoder-decoder-evaluator architecture. The encoder, denoted as \( g_{\text{enc}} \), maps feature transformation sequences \( s_i \in \mathcal{S} \) into a continuous embedding space \( \mathcal{E} \), such that:
\(
g_{\text{enc}}: \mathcal{S} \to \mathcal{E}.
\)
The decoder, denoted as \( g_{\text{dec}} \), reconstructs transformation sequences from the embeddings:
\(
g_{\text{dec}}: \mathcal{E} \to \mathcal{S}.
\)
The evaluator, denoted as \( h_{\text{eval}} \), estimates the performance of a model \( \hat{y} \) based on the embedding learned by the encoder, formalized as:
\(
h_{\text{eval}}: \mathcal{E}  \to \mathbb{R},
\)
where \( \mathbb{R} \) represents the space of performance metrics, such as accuracy or other evaluation measures.
The embedding space \( \mathcal{E} \) is constructed such that each embedding \( e_i \in \mathcal{E} \) corresponds to a specific feature transformation sequence \( s_i \) and its associated performance \( p_i \). To identify improved transformations, a gradient-based search is performed in the embedding space:
\(
e^* = \arg\max_{e \in \mathcal{E}} h_{\text{eval}}(g_{\text{dec}}(e)),
\)
where \( s^* = g_{\text{dec}}(e^*) \) represents the optimal transformation sequence found through this process.

This unified framework enables efficient exploration of the feature transformation landscape. By encoding feature learning knowledge into an embedding space, generative AI facilitates knowledge transfer across tasks and domains while optimizing transformations for better results. 
In the following sections, we will explore existing research on applying generative AI to feature selection and generation.

\subsubsection{Generative AI for Feature Selection} 
Pioneering studies have introduced a two-step generative approach to address inherent challenges in reinforcement learning-based feature selection, such as high-dimensional action spaces, redundancy, and limited generalization capabilities. These methods redefine feature selection as an optimization problem in a continuous embedding space, moving beyond the traditional discrete combinatorial framework. The first step involves collecting feature selection behaviors to construct a continuous hidden space. In the second step, selection decisions are conditionally generated based on this learned space, enabling more efficient and effective feature selection.

Among these approaches, GAINS~\cite{GAINS} marked a significant milestone by introducing a deep encoder-decoder-evaluator framework. This framework learns a discriminative embedding space to optimize feature subsets. By integrating reinforcement learning with classical selection methods for automated data preparation, GAINS ensures the diversity of training data and enhances the overall effectiveness of feature selection. Its ability to seamlessly combine generative modeling with classical techniques has laid the groundwork for further advancements.

Building on the encoder-decoder-evaluator paradigm, VTFS~\cite{VTFS} redefines feature selection as a sequential token generation task, offering a probabilistic perspective. Using a transformer-based variational autoencoder (VAE), VTFS embeds feature subsets into a continuous space. The self-attention mechanisms of the transformer effectively capture complex dependencies among features, while the VAE aligns the embedding distribution with a normal distribution. This alignment mitigates overfitting and reduces sensitivity to noise, ensuring robust and diverse feature subsets. VTFS demonstrates the potential of combining transformers with variational principles to achieve more adaptable feature selection.

Advancing this trajectory, FSNS~\cite{FNFS} addresses redundancy in high-dimensional feature spaces with an orthogonality-preserving embedding mechanism. Integrated into an encoder-decoder-evaluator framework, this mechanism minimizes redundant feature relationships within the embedding space, significantly improving the robustness of the selected subsets. Additionally, it facilitates the identification of more compact feature subsets without compromising predictive accuracy. By focusing on redundancy reduction, FSNS provides a novel solution to a long-standing challenge in feature selection.

Generative approaches have also shown their versatility in domain-specific applications. GERBIL~\cite{GERBIL} exemplifies this by tackling high-dimensional, low-sample-size (HDLSS) challenges prevalent in biomedical datasets. By embedding domain-specific biological knowledge into a continuous space, GERBIL enables the efficient identification of biomarker subsets critical for early disease detection, prognosis, and personalized medicine. This approach highlights the transformative potential of generative AI in healthcare, where precise feature selection can directly improve patient outcomes and reduce healthcare costs.

In summary, these generative AI methods—spanning encoder-decoder frameworks, variational embedding models, orthogonality-preserving mechanisms, and domain-specific adaptations—significantly enhance the effectiveness, stability, and generalization of feature selection. They together provide a robust foundation for future research and pave the way for broader applications in high-dimensional data environments.

\subsubsection{Generative AI for Feature Generation}
Inspired by the success of AutoML~\cite{karmaker2021automl} in automating machine learning tasks, generative AI has emerged as a transformative approach for feature generation. These advanced methods create optimal feature spaces by using sophisticated techniques in embedding, optimization, and generation. They typically sample transformed feature sets as observations, treating them as feature knowledge to construct a continuous embedding space, which is subsequently optimized to yield the most effective feature sets.

A notable early contribution is DIFER~\cite{zhu2022difer}, which introduced AutoML techniques into feature generation. DIFER uses randomly generated transformation-accuracy data to guide the feature generation process. However, this approach faces several limitations: the random generation of transformation-accuracy data often results in invalid training instances with inconsistent transformation performances; feature embeddings and reconstructions are handled independently, neglecting interactions between features; the manual specification of the number of generated features complicates the reconstruction process; and its greedy search strategy for transformation reconstruction can lead to suboptimal outcomes. These limitations highlight the need for more sophisticated methods to address the inherent challenges of feature generation.

Building on this foundation, Reinforcement-Enhanced Autoregressive Feature Transformation (MOAT)~\cite{NIPS@MOAT} advances the field by integrating reinforcement learning-based transformation sequence collection and redefining feature transformation as a continuous optimization problem. MOAT employs reinforcement learning to generate superior transformation sequences and utilizes gradient-directed search to identify optimal embeddings. Its innovative use of postfix expressions for compactly encoding transformation sequences enhances scalability and robustness in supervised tasks. However, MOAT’s reliance on the quality of downstream task predictors limits its applicability in unsupervised scenarios, highlighting the trade-offs between task-specific performance and generalization.

Addressing the dependence on labeled data, the Unsupervised Generative Feature Transformation Framework (NEAT)~\cite{KDD@NEAT} introduces a novel approach that eliminates the need for supervision. By employing graph-based contrastive learning and multi-objective fine-tuning, NEAT represents features as a similarity graph and optimizes feature sets through a differentiable embedding space. This framework excels in capturing non-linear feature interactions and enhancing generalization in unsupervised settings. However, NEAT’s effectiveness is contingent on the design and fine-tuning of robust graph augmentations and accurate unsupervised utility metrics, underscoring the complexity of fully autonomous feature generation.

More recently, advanced methodologies leveraging large language models (LLMs) have expanded the possibilities of feature generation. Zhang et al.~\cite{zhang2024dynamic} proposed a dynamic and adaptive automated feature generation process using LLMs. This approach restructures feature spaces based on downstream task feedback, utilizing the in-context learning and reasoning capabilities of LLMs to dynamically adapt and optimize generated features across a range of machine learning scenarios. Notably, it eliminates the need for additional machine learning models, streamlining the feature generation pipeline.

Extending the application of LLMs, Zhang et al.~\cite{zhang2024tifg} introduced Text-Informed Feature Generation (TIFG), which integrates textual information into feature generation through Retrieval-Augmented Generation (RAG) technology. By retrieving relevant features from external knowledge bases, TIFG generates explainable features that enrich the feature space and uncover feature relationships. This automated framework continuously optimizes the feature generation process, adapting to new data inputs and iteratively improving downstream task performance.

In the domain of feature weighting, Zhang et al.~\cite{zhang2024tfwt} addressed the limitations of traditional tabular data processing methods, which often assume equal importance across all features and samples. Their proposed Transformer-based approach captures complex feature dependencies and assigns contextual weights to discrete and continuous features, preserving critical feature information and enhancing model performance.

In summary, generative AI approaches for feature generation significantly enhance the creation of optimal feature spaces by constructing and optimizing continuous embedding spaces. While these methods address inefficiencies in discrete search spaces and improve the modeling of feature interactions, challenges remain. Issues such as handling invalid training data, capturing complex feature interactions, designing robust unsupervised utility metrics, and optimizing feature generation pipelines represent key areas for future research and improvement.

%% file: 6_comparative_analysis.tex
\section{Comparative Analysis of Feature Selection and Generation Approaches}

\subsection{Strengths and Limitations of Traditional vs. Advanced Methods}

Feature engineering, encompassing feature selection and generation, is foundational to building effective machine learning models. While traditional methods have long served as the cornerstone of this process, advanced paradigms like reinforcement learning (RL) and generative AI are reshaping the landscape of feature engineering. To better understand their roles, this section compares these two approaches across several critical dimensions, as summarized in Table \ref{tab: comparison}.

\begin{table*}[h]
\centering
\caption{Comparison of Traditional and Advanced Methods}\label{tab: comparison}
\begin{tabular}{c|c|c} 
\hline
Aspect           & Traditional Methods                                                                                          & Advanced Methods                                                                                      \\ 
\hline\hline
Performance      & \begin{tabular}[c]{@{}c@{}}Efficient for small datasets, \\struggles with high-dimensional data\end{tabular} & \begin{tabular}[c]{@{}c@{}}Scalable, handles complex patterns, \\but resource-intensive\end{tabular}  \\\hline
Interpretability & Highly interpretable                                                                                         & Requires additional tools for explainability                                                          \\\hline
Adaptability     & Suitable for static, structured data                                                                         & Handles multi-modal and dynamic datasets                                                              \\\hline
Automation       & Limited, requires manual tuning                                                                              & Highly automated and dynamic                                                                          \\
\hline
\end{tabular}
\end{table*}

\noindent\textbf{Performance and Scalability.}
Traditional methods are well-suited for small to medium-sized datasets, where computational efficiency and simplicity are key. Techniques such as filter-based statistical tests or wrapper methods like recursive feature elimination excel in manageable feature spaces, providing reliable results with minimal overhead. However, as datasets grow in scale or dimensionality, these methods face significant challenges. The computational cost of evaluating numerous feature subsets often becomes prohibitive, limiting their applicability in high-dimensional or large-scale contexts.
In contrast, advanced methods such as RL and generative AI are designed to scale effectively to handle high-dimensional datasets and model complex, non-linear interactions. By leveraging state-of-the-art optimization and embedding techniques, these methods can uncover intricate patterns that traditional approaches might miss. However, this scalability comes at a cost: advanced methods require substantial computational resources, including GPUs, and are associated with longer training times. These trade-offs highlight the need to balance performance gains with resource availability when selecting an approach.

\noindent\textbf{Interpretability.}
One of the defining strengths of traditional methods is their interpretability, making them particularly valuable in fields like finance and healthcare, where transparency is critical. Techniques such as statistical tests or regularization methods like Lasso~\cite{lasso1996} offer clear insights into feature importance, allowing domain experts to easily validate the results. But, the simplicity that underpins their interpretability also limits their capacity to capture non-linear relationships within the data.
In contrast, advanced methods often operate as black boxes, prioritizing performance and flexibility over interpretability. While RL and generative AI can reveal intricate patterns and generate novel features, understanding their decision-making processes frequently requires external tools like SHAP~\cite{NIPS@SHAP} or LIME~\cite{KDD@LIME}. These tools, while effective, add complexity to the workflow, creating an additional layer of effort for practitioners seeking both high performance and explainability.

\noindent\textbf{Adaptability and Automation.}
Adaptability and automation mark another area of distinction between traditional and advanced methods. Traditional methods rely heavily on domain expertise and manual intervention, making them well-suited for static and structured datasets. However, they struggle with multi-modal or dynamic data, often requiring extensive preprocessing to combine different data types, such as numerical and categorical features. This dependence on manual tuning limits their scalability in rapidly evolving or heterogeneous data environments.
Advanced methods, by contrast, are inherently more adaptable and automated. RL dynamically optimizes feature selection strategies, adjusting to changing data characteristics in real time. Similarly, generative AI learns latent representations to create new features that capture the underlying complexity of dynamic datasets. Despite these advantages, implementing and fine-tuning these methods can be resource-intensive and require advanced expertise, posing a barrier to adoption for teams with limited technical capabilities.

The comparison between traditional and advanced methods underscores the evolving nature of feature engineering. While traditional approaches offer simplicity, efficiency, and interpretability, they are constrained by their limited scalability and adaptability. Advanced methods, though resource-intensive and less interpretable, excel in handling high-dimensional, complex, and dynamic datasets, making them indispensable for modern machine learning applications. Understanding the trade-offs between these paradigms is crucial for selecting the most appropriate approach for a given problem.

\subsection{Guidelines and Best Practices}

Feature engineering is a cornerstone of tabular data-centric AI, refining and expanding datasets to improve downstream task performance. To ensure best practices, this section provides comprehensive guidelines for feature selection and generation, emphasizing simplicity, scalability, domain relevance, and ethical considerations.

\noindent \textbf{Optimize for Simplicity and Relevance.}
Feature engineering should begin with simplicity, focusing on reducing complexity while maximizing the utility of the data. Establishing a straightforward baseline provides a foundation for iterative improvements.
 \textit{Feature Selection:} Start by employing traditional methods such as filter or embedded techniques to identify relevant features. These methods strike a balance between computational efficiency and performance. For example, selecting features based on correlation or mutual information offers a practical starting point, avoiding unnecessary computational overhead while delivering meaningful results.
\textit{Feature Generation:} Initial feature generation efforts should leverage basic combinations of existing features, guided by domain knowledge to ensure relevance and interpretability. For instance, deriving a "profit margin" feature from revenue and cost data can provide actionable insights in a business context while maintaining simplicity.

\noindent \textbf{Balance Automation and Domain Knowledge.}
Effective feature engineering requires a balance between automation for scalability and domain knowledge for relevance and interpretability. These approaches complement one another when integrated into a cohesive workflow.
\textit{Harness Automation for Scalability and Efficiency:} Automated frameworks like MOAT~\cite{NIPS@MOAT} streamline feature transformation processes, optimizing high-dimensional feature spaces through continuous embedding. Such methods efficiently uncover complex interactions that are often infeasible to identify manually, making them invaluable for large-scale datasets.
\textit{Leverage Domain Knowledge for Relevance:} Domain expertise ensures that generated features align with real-world significance and remain interpretable. Experts can guide constraints on automated feature transformations to reflect meaningful relationships or validate that generated features meet industry-specific needs.
\textit{Integrate and Validate:} Combining automation with domain knowledge provides a scalable yet grounded approach. Iterative validation by domain experts ensures that automated outputs contribute meaningfully to the task and remain aligned with practical applications.

\noindent \textbf{Iterative Validation and Explainability.}
Feature engineering is a dynamic process that benefits from continuous refinement to enhance both model performance and feature interpretability.
\textit{Validation:} Regularly evaluate the impact of selected or generated features on downstream tasks using techniques such as cross-validation or feature importance rankings. As data evolves, reassessing feature contributions ensures that models remain effective.
\textit{Explainability:} Prioritize interpretability, particularly in sensitive fields like healthcare or finance, where transparency is critical. Tools like SHAP~\cite{NIPS@SHAP} can provide valuable insights into the influence of individual features on model predictions, fostering trust and accountability.

\noindent \textbf{Address Scalability and Resource Constraints.}
As datasets grow in size, scalability and computational efficiency become paramount in feature engineering.
\textit{Scalability:} Employ dimensionality reduction techniques such as PCA~\cite{PCA} or t-SNE~\cite{t-SNE} to condense high-dimensional data without significant loss of interpretability. Pair these methods with automated frameworks like deep feature synthesis to enhance scalability in handling large datasets.
\textit{Efficiency:} Optimize computational resources by beginning with simpler, more efficient methods during initial explorations. 
Reserve advanced, resource-intensive techniques for scenarios where traditional approaches fail to yield satisfactory results.

\noindent \textbf{Focus on Ethical and Practical Considerations.}
Feature engineering must not only prioritize technical efficiency but also adhere to ethical standards and practical constraints to ensure responsible implementation.
\textit{Ethical Compliance:} Avoid introducing biases or violating privacy regulations during feature engineering. Techniques like federated learning~\cite{FL} and differential privacy~\cite{DP} can protect sensitive data, ensuring compliance with ethical and legal standards.
\textit{Practical Implementation:} Maintain thorough documentation of all feature engineering steps to ensure reproducibility. Clearly communicate the rationale behind feature selection and generation to stakeholders, fostering trust, transparency, and alignment with project goals.

\subsection{When to Use RL and Generative AI for Feature Engineering}

Advanced methods such as RL and generative AI offer powerful tools for feature engineering, particularly when traditional approaches struggle to handle the complexity or scale of the problem. This section outlines scenarios where these methods are especially beneficial and provides practical considerations for their application.

\noindent\textbf{RL for Feature Engineering}
RL is well-suited for scenarios that require adaptive decision-making, dynamic data environments, or sequential optimization of feature sets.
\textit{Dynamic or Streaming Data:}
RL excels in environments where data evolves over time, such as streaming or time-series data. For instance, in real-time recommender systems~\cite{RecSys}, RL can dynamically select features that capture user behavior as it changes, ensuring models stay relevant.
\textit{Sequential Optimization Problems:}
When the feature engineering process involves multiple interdependent and dependent steps, such as iteratively selecting and refining features to maximize a downstream objective (e.g., accuracy, robustness, or interpretability), RL can efficiently explore the search space and learn optimal sequences of actions.
\textit{Complex Feature Interaction Discovery:}
RL is particularly effective in uncovering high-order feature interactions that would be computationally expensive to explore exhaustively through traditional methods. By leveraging its iterative framework, RL can dynamically identify and prioritize the most relevant combinations of features, enabling the discovery of intricate patterns and relationships.

\noindent\textbf{Generative AI for Feature Engineering.}
Generative AI can be a better choice for feature engineering when the task involves systematically exploring large, complex feature spaces and producing features with interpretable and traceable transformations. Unlike RL, which excels in sequential or adaptive decision-making, generative AI is particularly effective in scenarios requiring efficient, parallel exploration of the feature space or structured transformation of existing features.
\textit{Static or Semi-Static Data Environments:}
Generative AI is well-suited for datasets that are relatively stable or where the relationships between features do not evolve dynamically over time. It can systematically generate candidate features in one pass or through iterative refinement without requiring real-time adjustments.
\textit{Large Search Spaces with Complex Relationships:}
When the potential transformations or combinations of features form a large and complex search space, generative AI can leverage its modeling capabilities to explore these spaces more comprehensively. Unlike RL, which follows a step-by-step optimization process, generative models can simultaneously evaluate and synthesize multiple transformations, enabling faster convergence.
\textit{Batch Feature Generation for High-Dimensional Data:}
Generative AI is particularly efficient when a large number of candidate features need to be produced and evaluated in a batch. This is especially useful in high-dimensional datasets, where traditional or stepwise methods (like RL) may become computationally prohibitive.

%% file: 7_future_conclusion.tex
\section{Challenges and Future Research Directions}

As the field of tabular data-centric AI advances, several key challenges and opportunities for innovation have emerged. 
This section explores these challenges and outlines future research directions, focusing on automation, interpretability, privacy, scalability, and the integration of cutting-edge technologies such as LLMs and multimodal systems.

\subsection{Emerging Trends in Tabular Data-Centric AI}

\noindent\textbf{Automation in Feature Engineering.}
The automation of feature engineering has revolutionized the handling of tabular data, reducing manual effort and enabling efficient workflows. 
However, the balance between generalization and domain-specific customization remains a significant hurdle. 
While AutoML methods excel at providing baseline transformations, they often fail to incorporate nuanced domain knowledge crucial for real-world applications. Additionally, their scalability is constrained when applied to large-scale, high-dimensional datasets, limiting their utility in dynamic and resource-constrained environments.
Future research should emphasize the development of hybrid systems that integrate human-in-the-loop frameworks with automated processes. These systems would combine the adaptability of machine learning with domain expertise, ensuring both efficiency and relevance. Moreover, optimizing resource consumption for these methods will be critical, enabling their deployment in environments with limited computational capacity and ensuring broader accessibility.

\noindent\textbf{Explainable AI and Interpretability.}
Feature engineering plays a pivotal role in determining model performance, yet it is often treated as a "black box" process. In high-stakes domains such as healthcare, finance, and legal analytics, understanding the rationale behind feature creation is as important as the model predictions themselves. While explainability tools for model outputs are well-established, their application to feature engineering remains underexplored.
To bridge this gap, future solutions should prioritize designing transformation workflows that are inherently interpretable. Visual and metric-based explanations can serve as a bridge between technical practitioners and non-technical stakeholders, fostering transparency. Additionally, methods to quantify the impact of specific feature transformations on model performance will enable a deeper understanding of their contributions, promoting trust and accountability in feature engineering pipelines.

\noindent\textbf{Privacy-Conscious Feature Engineering with Federated Learning.}
The growing emphasis on data privacy has brought federated learning to the forefront as a framework for collaborative feature engineering across distributed datasets. However, implementing privacy-conscious transformations presents unique challenges, including handling heterogeneous data distributions, minimizing communication overhead during iterative learning, and aligning feature semantics across participants.
Future research should address these challenges by developing robust feature alignment and optimization techniques that respect privacy constraints. Enhancing encryption-enabled federated systems can ensure data security without compromising performance. Furthermore, designing scalable architectures capable of handling both vertical and horizontal data partitioning efficiently will be instrumental in advancing privacy-conscious feature engineering.

\noindent\textbf{LLMs and Multimodality for Feature Generation.}
Large Language Models (LLMs) and multimodal systems are opening new frontiers in feature generation by processing diverse data types, including text, images, and structured tabular data. These systems hold immense potential for cross-domain feature engineering. However, challenges persist in encoding tabular data effectively within LLM architectures, which are traditionally optimized for textual inputs. Additionally, integrating multimodal data while preserving the unique characteristics of each modality remains a complex task.
Future efforts should focus on designing advanced embedding techniques specifically tailored to tabular data, enabling its seamless integration into LLM frameworks. Methods for aligning features across modalities must ensure that semantic relationships are preserved, thereby enhancing the utility of multimodal representations. Efficient fine-tuning and knowledge transfer mechanisms for tabular tasks will further expand the practicality and adoption of LLM-based feature generation systems.

\subsection{Open Challenges in Future Research}

\noindent\textbf{Scalability and Efficiency of Advanced Methods.}
As datasets grow in complexity and scale, ensuring the efficiency of future methods is increasingly challenging. 
Existing approaches often require substantial computational resources, making them impractical for real-time or resource-constrained applications. The iterative nature of feature generation and evaluation further compounds these bottlenecks, posing significant barriers to scalability.
Addressing this challenge calls for the development of lightweight algorithms optimized for distributed and cloud-based infrastructures. Leveraging approximate methods, parallel processing, and hardware acceleration can dramatically reduce computational overhead while maintaining transformation quality. These advancements will be essential to making advanced methods more accessible and deployable in both academic research and industrial applications.

\noindent\textbf{Interpretability and Explainability in Feature Engineering.}
As feature engineering methodologies grow more sophisticated, ensuring their interpretability remains a persistent challenge. Complex transformation pipelines can obscure the relationships between raw inputs and model outputs, creating barriers to trust and adoption—particularly in sensitive domains where accountability is paramount.
Future research should aim to create frameworks that balance performance with transparency. Designing transformation pipelines with traceable and interpretable workflows will help users understand the logic and impact of feature engineering processes. User-friendly tools for visualizing transformation steps and feature contributions can further empower stakeholders, bridging the gap between machine intelligence and human comprehension.

\section{Conclusion Remarks}
Tabular data remains a cornerstone of AI, driving critical applications across diverse domains. This survey has provided a comprehensive overview of tabular data-centric AI, tracing its evolution from traditional methods to advanced approaches like reinforcement learning and generative AI. By addressing challenges such as scalability, interpretability, and privacy, we have highlighted both the limitations and transformative potential of these methods.
Emerging trends, including AutoML automation, LLM integration, and multimodal systems, herald a promising future for feature engineering. These advancements bridge the gap between domain expertise and automated solutions, enabling scalable, interpretable, and efficient workflows that can tackle increasingly complex datasets.
We underscore the importance of hybrid approaches that combine the precision of human expertise with the efficiency of machine learning. As research progresses to address open challenges, tabular data-centric AI is positioned to unlock new opportunities, fostering innovation and expanding its influence across industries.
This survey aims to inspire continued exploration and innovation, empowering researchers and practitioners to fully harness the untapped potential of tabular data in advancing AI and its revolutionized applications.